\def\paperID{6923}
\def\confName{CVPR}
\def\confYear{2025}

\def\paperTitle{Towards Unified Molecule-Enhanced Pathology Image Representation Learning via Integrating Spatial Transcriptomics}

\author{
Minghao~Han$\textsuperscript{1,2}$$\quad$ 
Dingkang~Yang$\textsuperscript{1,2}$\footnotemark[4]$\quad$ 
Jiabei~Cheng$\textsuperscript{3}$$\quad$ 
Xukun~Zhang$\textsuperscript{1,2}$$\quad$ \\
Linhao~Qu$\textsuperscript{4}$$\quad$ 
Zizhi~Chen$\textsuperscript{1,2}$$\quad$ 
Lihua~Zhang$\textsuperscript{1,2}\footnotemark[4]$
\\
\\
$\textsuperscript{1}$Academy for Engineering and Technology, Fudan University, Shanghai, China \\
$\textsuperscript{2}$Cognition and Intelligent Technology Laboratory, Fudan University, Shanghai, China \\
$\textsuperscript{3}$Department of Automation, Shanghai Jiaotong University, Shanghai, China \\
$\textsuperscript{4}$Fudan University, Shanghai, China \\
{\tt\small mhhan22@m.fudan.edu.cn, dkyang20@fudan.edu.cn, lihuazhang@fudan.edu.cn}
}

\newif\ifreview 
\newif\ifarxiv \newcommand{\arxiv}{\arxivtrue}
\newif\ifcamera 
\newif\ifrebuttal 
\newcommand{\ours}{\textsc{Umpire}}
\newcommand{\ie}{{\emph{i.e.}},\xspace}

\arxiv 

\pdfoutput=1
\documentclass[10pt,twocolumn,letterpaper]{article}
\ifreview \usepackage[review]{cvpr} \fi
\ifarxiv \usepackage[pagenumbers]{cvpr} \fi
\ifrebuttal \usepackage[rebuttal]{cvpr} \fi
\ifcamera \usepackage{cvpr} \fi

\usepackage{graphicx}	
\usepackage{amsmath}	
\usepackage{amssymb}	
\usepackage{booktabs}
\usepackage{times}
\usepackage{microtype}
\usepackage{epsfig}
\usepackage[table,xcdraw,dvipsnames]{xcolor}
\usepackage{caption}
\usepackage{float}
\usepackage{placeins}
\usepackage{color, colortbl}
\usepackage{stfloats}
\usepackage{enumitem}
\usepackage{tabularx}
\usepackage{xstring}
\usepackage{multirow}
\usepackage{xspace}
\usepackage{url}
\usepackage{subcaption}
\usepackage{xcolor}
\usepackage[hang,flushmargin]{footmisc}

\ifcamera \usepackage[accsupp]{axessibility} \fi

% ...

\ifarxiv  \fi

\newcommand{\R}[1]{{%
    \textbf{%
        \ifstrequal{#1}{1}{\textcolor{red}{R#1}}{%
        \ifstrequal{#1}{2}{\textcolor{blue}{R#1}}{%
        \ifstrequal{#1}{3}{\textcolor{magenta}{R#1}}{%
        \ifstrequal{#1}{4}{\textcolor{teal}{R#1}}{%
                           \textcolor{cyan}{R#1}%
        }}}}%
    }%
}}

\usepackage{xr-hyper}

\makeatletter
\newcommand*{\addFileDependency}[1]{
  \typeout{(#1)}
  \@addtofilelist{#1}
  \IfFileExists{#1}{}{\typeout{No file #1.}}
}

\makeatother

\definecolor{cvprblue}{rgb}{0.21,0.49,0.74}
\usepackage[pagebackref,breaklinks,colorlinks,citecolor=cvprblue]{hyperref}
\usepackage[capitalize]{cleveref}
\crefname{section}{Sec.}{Secs.}
\crefname{table}{Table}{Tables}
\crefname{figure}{Fig.}{Figs.}

\ifarxiv \crefname{appendix}{App.}{Apps.}
\else \crefname{appendix}{Suppl.}{Suppls.} \fi

\usepackage{makecell}
\usepackage{graphicx, diagbox}
\usepackage[ruled,linesnumbered]{algorithm2e}
\usepackage{times}

\begin{document}
\title{\paperTitle}
\maketitle
\renewcommand{\thefootnote}{\fnsymbol{footnote}} 
\footnotetext[4]{Corresponding authors.} 
\begin{abstract}
Recent advancements in multimodal pre-training models have significantly advanced computational pathology. However, current approaches predominantly rely on visual-language models, which may impose limitations from a molecular perspective and lead to performance bottlenecks. Here, we introduce a \textbf{U}nified \textbf{M}olecule-enhanced \textbf{P}athology \textbf{I}mage \textbf{RE}presentationn Learning framework ($\ours$). $\ours$ aims to leverage complementary information from gene expression profiles to guide the multimodal pre-training, enhancing the molecular awareness of pathology image representation learning. We demonstrate that this molecular perspective provides a robust, task-agnostic training signal for learning pathology image embeddings. Due to the scarcity of paired data, approximately 4 million entries of spatial transcriptomics gene expression were collected to train the gene encoder. By leveraging powerful pre-trained encoders, $\ours$ aligns the encoders across over 697K pathology image-gene expression pairs. The performance of $\ours$ is demonstrated across various molecular-related downstream tasks, including gene expression prediction, spot classification, and mutation state prediction in whole slide images. Our findings highlight the effectiveness of multimodal data integration and open new avenues for exploring computational pathology enhanced by molecular perspectives. The code and pre-trained weights are available at \url{https://github.com/Hanminghao/UMPIRE}.
\end{abstract}

\vspace{-0.2cm}
\section{Introduction}
\label{sec:intro}

Whole slide images (WSIs) and pathology images are considered the ``gold standard" for cancer analysis due to their capacity to provide detailed information at cellular and tissue levels~\cite{glaser2017light, ludwig2005biomarkers}. Recent advancements in Computational Pathology (CPATH) have leveraged deep learning to achieve significant progress in various tasks, including cancer diagnosis~\cite{dsmil,transmil, liu2024exploiting}, survival analysis~\cite{song2024multimodal, jaume2023modeling, wang2024pathology}, and cancer staging~\cite{wang2024mgiml, shi2023structure}. However, most existing paradigms focus on specific tasks and train models in isolation, which can cause these meticulously designed models to fail when faced with new data or tasks requiring retraining. Some researchers argue that instead of investing considerable effort in designing complex downstream models, it is more cost-effective and scientifically sound to develop foundational models that can adapt to a wide range of downstream tasks~\cite{chen2024uni,filiot2023phikon,hoptimus,xu2024whole,conch,plip}.

Recent research has demonstrated that utilizing a large number of noisy image-text pairs for extensive multimodal pre-training can enhance the alignment of spatial representations between images and text, as well as improve the encoder's performance on downstream tasks~\cite{clip, li2022blip, align, yu2022coca}. Building on this idea, several researchers have proposed contrastive learning-based pre-training frameworks that leverage pathology images and descriptive texts, including PLIP~\cite{plip} and CONCH~\cite{conch}. Despite the widespread of natural language in cancer pathology analysis, multimodal pre-training of image-text pairs fails to provide additional insights for cancer analysis. In contrast, gene expression data, such as RNA transcriptome, provides complementary information at the molecular level, elucidating the mechanisms of oncogenesis and facilitating personalized treatment recommendations~\cite{xu2024multimodal, ding2023pathology}. Consequently, TANGLE~\cite{tangle} introduced a methodology that employs bulk RNA to guide WSI representation learning. Their experimental results indicate that pre-training based on WSI and bulk RNA significantly enhances model performance on cancer subtype classification. However, their approach relies on WSIs and bulk RNA, representing only patient-level information and failing to capture the inherent heterogeneity within individual samples~\cite{li2021bulk,oksza2024caclust}.

Spatial Transcriptomics (ST) is an emerging technique that integrates pathology slides with gene expression (RNA transcriptome) analysis, enabling researchers to localize and quantify RNA expression within tissues~\cite{jain2024spatial}. In recent years, various ST methodologies, such as Spatial Transcriptomics~\cite{staahl2016visualization}, Visium~\cite{visium}, MERFISH~\cite{merfish}, and Xenium~\cite{xenium}, have advanced rapidly, establishing themselves as crucial links between pathology images and gene expression. Similar to image-text pairs, ST generates numerous mappings between pathology images and gene expression. Under typical conditions, pathology images specialize in the analysis of tissue structures and cell morphology~\cite{qu2024rethinking, shi2023structure}, while gene expression profiles excel in analyzing the tumor microenvironment and disease mechanisms~\cite{schaar2024nicheformer, elhanani2023spatial}. Both are crucial for cancer analysis. Recently, there has been rapid progress in the research of foundational and pre-trained models in both fields~\cite{chen2024uni, filiot2023phikon, hoptimus, cui2024scgpt,schaar2024nicheformer,geneformer}. However, a unified pre-training framework that integrates them is still lacking, leading to an incomplete perspective. This is due to two main factors: 1) Pathology images and gene expression data often originate from different labs and clinical environments, with varying formats and standards, which limits the construction of large-scale datasets; 2) Despite advances in visual-language models, there is no effective cross-modality learning framework for integrating pathology images with gene expression.

To address these challenges, we propose a two-stage \textbf{U}nified \textbf{M}olecule-enhanced \textbf{P}athology \textbf{I}mage \textbf{RE}presentationn Learning framework, termed $\ours$. It is well established that gene expression plays a crucial role in regulating cellular proliferation and intercellular interactions~\cite{zhu2019metabolic}. Anomalies in gene expression correspond to discernible morphological patterns in pathology images~\cite{kueckelhaus2024inferring}. Accordingly, we believe that leveraging gene expression to guide the representation learning of pathology images provides a more robust training signal than relying on image augmentation or text descriptions, enhancing the molecular perspective in this learning process. To our knowledge, $\ours$ is the first large-scale pre-training of pathology images and ST gene expression, providing a foundation for subsequent molecular perception pathological representation learning and multimodal integration models.

In this work, approximately 4M entries of ST gene expression were initially collected to pre-train a BERT-like gene encoder~\cite{bert}. Then, we filtered data from the HEST dataset to obtain 697K aligned pairs. Following established multimodal contrastive learning paradigms~\cite{clip,align,yu2022coca}, we aligned the vision encoder with the gene encoder during the alignment phase. Ultimately, extensive evaluations were conducted across multiple tasks, including bimodal gene expression prediction, unimodal spot/patch classification, and mutation state prediction for WSIs. Experimental results demonstrate that $\ours$ outperforms the baseline across all tasks. We also conducted comprehensive ablation experiments, visualization analyses, and case studies.

\section{Related Work}
\label{sec:related}
\vspace{0.5pt}\noindent\textbf{Self-supervised Representation Learning: }
By generating its own supervisory signals, self-supervised learning (SSL) can operate without manual labels. This approach has gained significant attention in recent research~\cite{mocov3,he2022masked}. SSL gained popularity in natural language processing (NLP) with models such as GPT~\cite{gpt3} and BERT~\cite{bert}, which employed SSL to learn semantic representations from text through tasks like masked language modeling. Due to similarities such as discrete sequences and context dependence, many NLP SSL techniques have been adapted for single-cell representation learning~\cite{cui2024scgpt,geneformer}. SSL has also gained traction in computer vision, with methods such as SimCLR~\cite{simclr} and MoCo~\cite{mocov3} learning visual representations through augmented views. This paper employs a BERT-like architecture to pre-train a gene encoder, which is then integrated into a multimodal contrastive learning framework.

\vspace{0.5pt}\noindent\textbf{Contrastive Learning: }
Contrastive learning is a powerful pre-training technique in the domain of SSL used to acquire task-agnostic representations. This mechanism constructs paired samples to enhance the proximity of paired embeddings in the latent space while increasing the distance between unpaired embeddings. PLIP~\cite{plip} collected 208K pairs of pathology images and captions from Twitter and fine-tuned the model based on CLIP, resulting in an encoder exhibiting robust performance across various downstream tasks. CONCH~\cite{conch} used over 1.17 million pathology image-caption pairs for task-agnostic pre-training, achieving excellent performance across 14 downstream benchmarks while minimizing the need for supervised fine-tuning. TANGLE~\cite{tangle} enhances performance on WSI level visual recognition tasks by aligning expression profiles with slide representations. Our $\ours$ aligns with these concepts by correlating pathology images with gene expression.

\vspace{0.5pt}\noindent\textbf{Spatial Transcriptomics in CPATH: }
Gene expression profiles offer a molecular perspective that complements tissue pathology, enabling researchers to better understand cancer pathogenesis and develop personalized treatment strategies~\cite{xu2024multimodal, ding2023pathology}. However, acquiring detailed gene expression profiles is time-consuming and costly~\cite{tian2023expanding,jain2024spatial}. Given the mapping between pathology images and gene expression profiles, some researchers have proposed predicting gene expression from pathology images~\cite{stnet,histogene,his2st,bleep,mclstExp}. BLEEP~\cite{bleep} and mclSTExp~\cite{mclstExp} employ contrastive learning to create a low-dimensional joint embedding space, enabling the estimation of gene expression in any pathology image patch using expression profiles from a reference dataset. However, these methods depend on training from scratch with a single dataset, leading to suboptimal model performance due to limited training data. We recommend pre-training encoders on large-scale datasets and fine-tuning on downstream tasks, as this approach improves performance and reduces computational costs compared to existing methods.

\section{Methodology}
\label{sec:method}

\begin{figure*}[tp]
    \centering
    \includegraphics[width=1\linewidth]{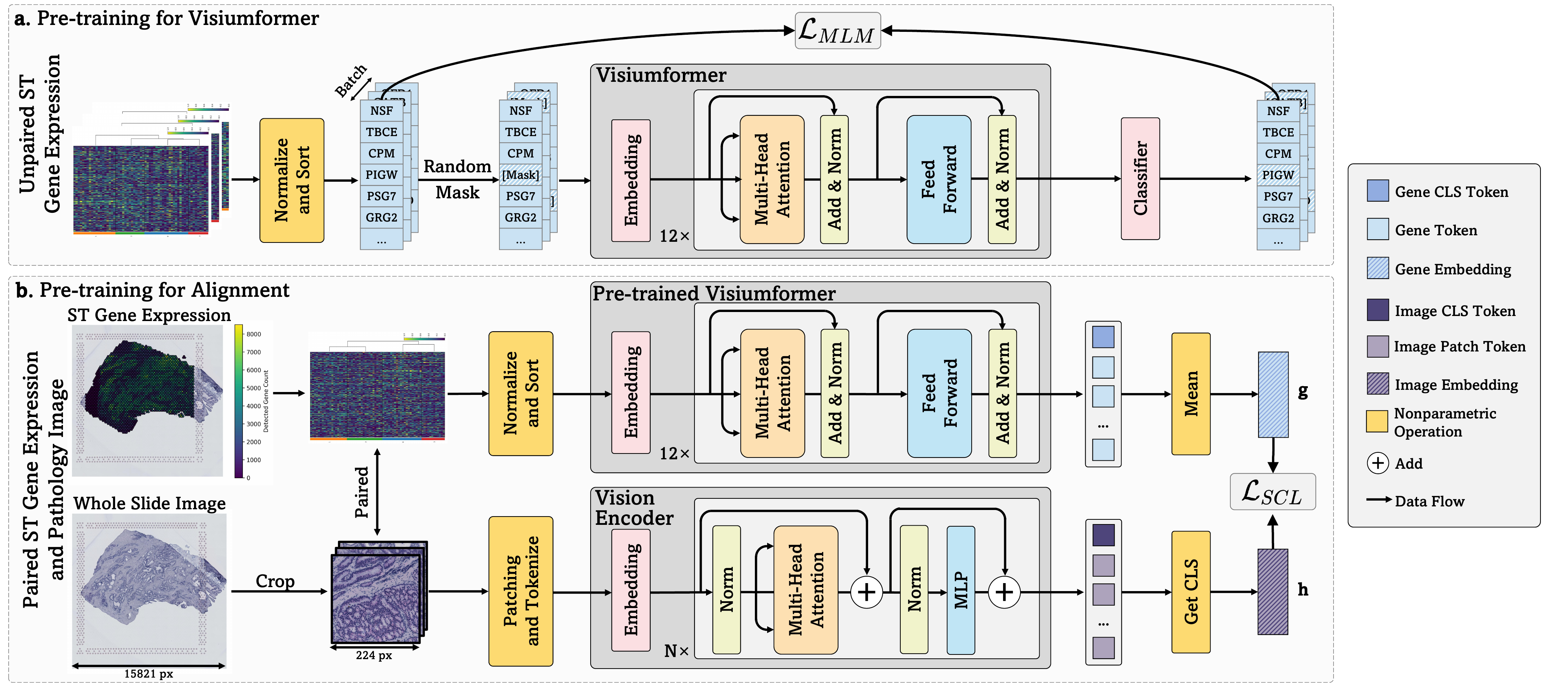}
    \caption{
    \textbf{Overview of $\ours$.} 
    First, approximately 4 million unlabeled spatial transcriptomics (ST) gene expression data were used to pre-train the Visiumformer for gene encoding. Next, a pre-trained pathologic Vision Transformer was adopted as the vision encoder. The symmetric contrastive loss \( \mathcal{L}_{\text{SCL}} \) is applied to align embeddings from both modalities.}
    \label{fig:fig1}
\end{figure*}

\subsection{Data Collection}
Given 1) the substantial heterogeneity of data across various sequencing platforms~\cite{schaar2024nicheformer}, 2) the 55-micron resolution of Visium, which aligns with the dimensions of individual tissue patches~\cite{jain2024spatial}, 3) the wider variety of genes detectable by Visium~\cite{jain2024spatial}, and 4) the relatively abundant and readily accessible Visium datasets~\cite{hest,chen2024stimage}, only Visium spatial transcriptomics (ST) gene expression was selected for training. Despite being solely pre-trained on Visium data, our model successfully demonstrates transferability and generalization to other sequencing platforms (Section \ref{sec:4.3}).

For gene expression, we collected approximately 4M ST gene expression data points from the Gene Expression Omnibus (GEO) and other public datasets~\cite{hest,chen2024stimage,yuan2023sodb,xu2024stomicsdb}. To our knowledge, this dataset represents the largest \textbf{Vi}sium-based \textbf{S}patial \textbf{T}ranscript\textbf{Omics} Dataset (ViSTomics-4M), encompassing 3.94 million ST data points collected from 1,363 slides across 180 datasets and publications. For further details about ViSTomics-4M, please refer to \textbf{Appendix} \ref{D.1}. For paired data, it was sourced from the largest pathology image and ST dataset, HEST~\cite{hest}. After filtering for human samples based on Visium, 697K aligned pathology image-gene expression pairs were obtained.

\subsection{Unsupervised Training for Unimodal Encoder}
Although HEST is the largest dataset in the field, it contains only 329 slides and 697K data pairs after filtering, which is still insufficient compared to other multimodal pre-training models (e.g., CLIP~\cite{clip} with 400M pairs and CONCH~\cite{conch} with 1.17M pairs). We initialize the encoders with pre-trained weights and subsequently align them in the latent space to address this limitation. While existing models for gene expression primarily focus on single-cell~\cite{cui2024scgpt, geneformer} or single-cell-level ST~\cite{schaar2024nicheformer}, ViSTomics-4M was collected to pre-train the gene encoder. Specifically, as shown in Figure \ref{fig:fig1}\textcolor{red}{a}, we developed a Transformer-based gene encoder, termed Visiumformer. For comparison, Nicheformer~\cite{schaar2024nicheformer} was also used as the gene encoder, though it focuses on single-cell ST data and has not been trained on Visium-based data.

\vspace{0.5pt}\noindent\textbf{Visiumformer Tokenization.}\hspace{1ex} We adopted a vocabulary including 20,310 genes. The average expression level for each gene across all samples was first calculated. To reduce batch effects, each gene expression value was normalized by dividing it by the average expression of the corresponding gene. Since each sequencing dataset originates from a whole tissue section, the data lacks an inherent order, rendering it order-agnostic~\cite{visium}. Therefore, we normalized the gene expression values and sorted them in descending order for each gene to complete the tokenization process:
\begin{small} 
\begin{equation}
\begin{split}
T_{i} & = \bigl\{ id(ep_{i}^{0}), id(ep_{i}^{1}), \ldots, id(ep_{i}^{n}): ep_{i}^{k}\ge ep_{i}^{k+1}\bigr\},
\end{split}
\end{equation}
\end{small}%
where $id(ep_{i}^{k})$ and $ep_{i}^{k}$ represent the index of gene $k$ in the gene vocabulary and the normalized gene expression of sample $i$. In this study, we set $ n $ to 1500, meaning that the context length for the gene encoder is 1500 tokens.

\vspace{0.5pt}\noindent\textbf{Visiumformer Pre-training.}\hspace{1ex} Given a tokenized ST gene expression $ T_{i} \in \mathbb{R}^{N} $, Visiumformer first applies an embedding process:
\begin{small} 
\begin{align}
x_{i}=Embedding(T_i{})+PosEmbedding(Pos_{i}),
\end{align}
\end{small}%
where $x_i \in \mathbb{R}^{N \times D}$ represents the vector to be fed into the Transformer block, $D$ is the input dimension, and $Pos_{i}=\{0,1,...,N-1\}$. Visiumformer is composed of 12 stacked Transformer blocks. Given the embedded sequence $x_i \in \mathbb{R}^{N \times D}$, each Transformer block processes the input sequence according to the following equations: 
\begin{small} 
\begin{gather}
x_{i}^{0} = x_{i}, \\
x_{i}^{l+1} = TransformerBlock(x_{i}^{l}).
\end{gather}
\end{small}%
In line with BERT~\cite{bert}, masked language modeling (MLM) loss is utilized to optimize Visiumformer. Specifically, $15\%$ of the tokens are randomly masked, and the model is trained to predict these masked tokens using the unmasked tokens as context. The MLM loss can be expressed as:
\begin{small} 
\begin{align}
\mathcal{L}_{MLM}=-\frac{1}{\left |M  \right | } {\textstyle \sum_{j\in M}^{}logP(t _{i,j}|T_{i})},
\end{align}
\end{small}%
where $M$ is the set of masked tokens, $T_i$ are the input tokens and $t_{i,j}$ is the $j$-th masked token of $T_i$.

\vspace{0.5pt}\noindent\textbf{Vision Encoder.}\hspace{1ex} The development of pathological visual foundation models has progressed rapidly~\cite{xu2024whole, vorontsov2023virchow, azizi2023robust, wang2022transformer, filiot2023phikon, chen2024uni}. In this study, we select Phikon (ViT-B/16, 86M)~\cite{filiot2023phikon} and UNI (ViT-L/16, 307M)~\cite{chen2024uni} as our vision encoders.

\begin{figure*}[tp]
    \centering
    \includegraphics[width=1\linewidth]{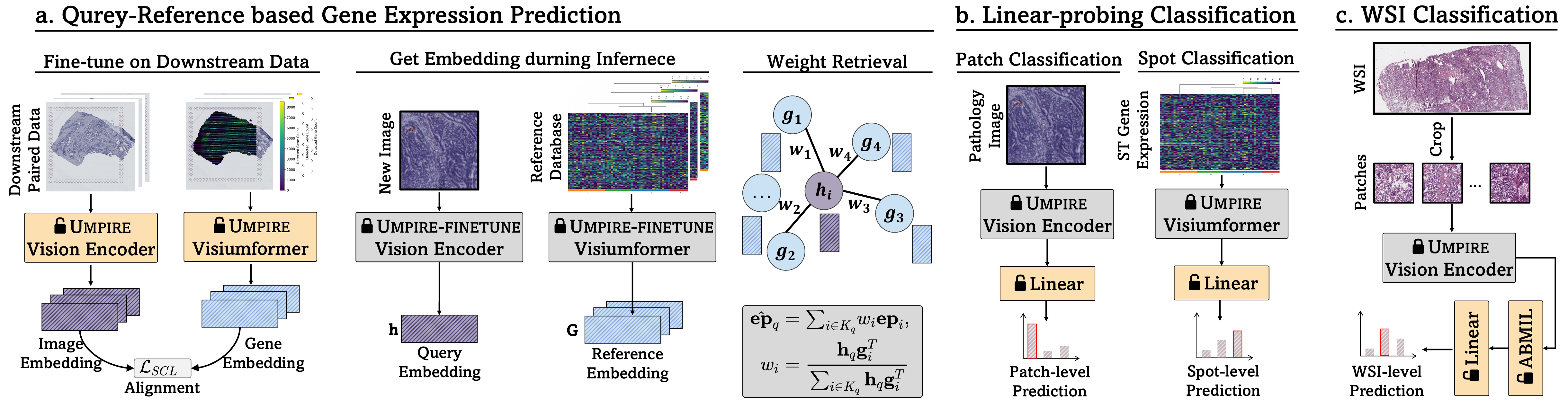}
    \caption{
    \textbf{Evaluation of Downstream Tasks.} 
    $\ours$ and baselines are assessed on: \textbf{a.} Bimodal gene expression prediction; \textbf{b.} Unimodal patch/spot classification; \textbf{c.} Vision-based WSI mutation state prediction.}
    \label{fig:fig2}
\end{figure*}

\subsection{Multimodal Alignment}
\label{sec:method:alignment}
\textbf{Cross-modality Alignment.}\hspace{1ex} 
As depicted in Figure \ref{fig:fig1}\textcolor{red}{b}, symmetric contrastive learning (SCL) loss was employed to align image embeddings with gene embeddings. Specifically, for a batch of $ M $ paired pathology image-gene expression samples $\{(\mathbf{h}_i, \mathbf{g}_i)\}_{i=1}^M$, where $\mathbf{h}_i$ and $\mathbf{g}_i$ denote the $ i $-th image and gene embedding obtained from the encoders, the loss function is defined as:
\begin{small}
\begin{equation}
\begin{aligned}[b]
\mathcal{L}_{SCL} &= -\frac{1}{2M} \sum_{i=1}^{M} \log \frac{\exp(\tau \mathbf{h}_i^T \mathbf{g}_i)}{\sum_{j=1}^{M} \exp(\tau \mathbf{h}_i^T \mathbf{g}_j)} \\
                  &\quad -\frac{1}{2M} \sum_{n=1}^{M} \log \frac{\exp(\tau \mathbf{g}_n^T \mathbf{h}_n)}{\sum_{m=1}^{M} \exp(\tau \mathbf{g}_n^T \mathbf{h}_m)},
\end{aligned}
\end{equation}
\end{small}%
where \(\tau\) is the temperature parameter. The first term represents image-to-gene loss, and the second represents gene-to-image loss. The loss function \(\mathcal{L}_{SCL}\) aims to minimize the distance between paired embeddings while maximizing the distance between unpaired embeddings.

\vspace{0.5pt}\noindent\textbf{Other Optimization Strategy.}\hspace{1ex} Unlike qualitative text, gene expression is quantitative, prompting us to consider a regression approach for aligning the encoders across modalities. As a complement to the primary method, a reconstruction loss (mean squared error) is introduced, termed 
\begin{small}
$\ours$-\textsc{Rec}:
\begin{align}
\mathcal{L}_{REC}=\frac{1}{M} {\textstyle \sum_{i=1}^{M}} \left \|\mathbf{ep}_i^{hvg} -MLP(\mathbf{h}_i) \right \|  _2,
\end{align}
\end{small}%
where $\mathbf{ep}_i^{hvg}$ represents the normalized top 1500 highly variable gene expression and $\mathbf{h}_i$ denotes the image embedding. Additionally, we employed various contrastive learning loss and L1 loss for ablation studies (in Section \ref{sec:4.4}).

\subsection{Query-Reference for Expression Prediction}
When attempting to learn full-dimensional gene expression from pathology images, regression-based approaches may struggle due to the ``curse of dimensionality''~\cite{bleep,mclstExp}. We mitigate this issue by fine-tuning, querying, and weighted aggregation (Figure \ref{fig:fig2}\textcolor{red}{a}). Specifically, $\ours$ first undergoes fine-tuning on the downstream dataset. During inference, the frozen vision encoder converts pathology images into query vectors \(\mathbf{h} \in \mathbb{R}^{Q \times d}\). Concurrently, all gene expression from the training set (termed reference database) is encoded into reference vectors \(\mathbf{g} \in \mathbb{R}^{R \times d}\) using the frozen gene encoder. The cosine similarity between the query and reference vectors is then computed. Finally, the top $K$ references for each query are identified, and a weighted method is applied to derive the predicted gene expression:
\begin{small}
\begin{align}
\hat{\mathbf{ep}}_q &=  {\textstyle \sum_{i \in K_q}^{}} w_i \mathbf{ep}_i,\\
w_i &=\frac{ \mathbf{h}_q\mathbf{g}_i^T}{\sum_{i \in K_q} \mathbf{h}_q\mathbf{g}_i^T}.
\end{align}
\end{small}%
\(\hat{\mathbf{ep}}_q\) represents the predicted gene expression associated with the query image \(q\), while \(K_q\) denotes the set of the top \(K\) nearest references for this query. Additionally, \(\mathbf{ep}_i\) signifies the authentic gene expression linked to reference \(i\).

\section{Experiments and Results}
\label{sec:experiments}

\subsection{Pre-training Implementation Details}
We first conducted vocabulary masking pre-training\cite{bert} of Visiumformer on ViSTomics-4M, with the entire training process spanning 1 million steps and a global batch size of 256. For the vision encoder, two pathology-specific vision encoders were selected: Phikon (ViT-B/16, 86M)~\cite{filiot2023phikon} and UNI (ViT-L/16, 307M)~\cite{chen2024uni}. A linear projection head was employed to map the image and gene embeddings into a 512-dimensional latent space for alignment. Each $\ours$ model under different combinations was trained for ten epochs with a global batch size of 512 during alignment. All pre-training tasks were performed on four NVIDIA A800 GPUs. Please refer to \textbf{Appendix} \ref{B.2} for details.

\begin{table*}
\centering
\resizebox{\textwidth}{!}{
\begin{tabular}{c|l|cccccc|c} 
\toprule
\multirow{2}{*}{Top 50}                                                            &\multicolumn{1}{c}{\multirow{2}{*}{Method}}  & \multicolumn{2}{c}{\textbf{HLT}}          & \multicolumn{2}{c}{\textbf{HPC}}          & \multicolumn{2}{c|}{\textbf{HER2+}} & \multirow{2}{*}{\textbf{Average}}  \\
                                                                                  &                         & HVG   & \multicolumn{1}{c|}{HEG} & HVG   & \multicolumn{1}{c|}{HEG} & HVG   & HEG                &                           \\ 
\midrule
\multirow{4}{*}{\begin{tabular}[c]{@{}c@{}}Regression\\based\end{tabular}}        & ST-Net~\cite{stnet}       & 0.0421$_{\pm0.0206}$  & 0.0406$_{\pm0.0140}$  & 0.2172$_{\pm0.1720}$    & 0.0445$_{\pm0.0386}$  & 0.1129$_{\pm0.0576}$  & 0.0940$_{\pm0.0421}$  & 0.0919                      \\
                                                                                  & HisToGene~\cite{histogene}& 0.0357$_{\pm0.0213}$  & 0.0414$_{\pm0.0322}$  & 0.1338$_{\pm0.1093}$    & 0.0912$_{\pm0.0451}$  & 0.0329$_{\pm0.0416}$  & 0.0287$_{\pm0.0387}$  & 0.0606                      \\
                                                                                  & His2ST~\cite{his2st}      & 0.0054$_{\pm0.0122}$  & 0.0029$_{\pm0.0163}$  & 0.0252$_{\pm0.0213}$    & 0.0127$_{\pm0.009}$   & 0.0443$_{\pm0.0197}$  & 0.0328$_{\pm0.0174}$  & 0.0206                       \\
                                                                                  & THItoGene~\cite{thitogene}& 0.0063$_{\pm0.0098}$  & 0.0020$_{\pm0.0106}$  & 0.0294$_{\pm0.0316}$    & 0.0163$_{\pm0.094}$   & 0.0391$_{\pm0.0146}$  & 0.0286$_{\pm0.0167}$  & 0.0203                       \\ 
\midrule
\multirow{2}{*}{\begin{tabular}[c]{@{}c@{}}Contrastive learning\\based\end{tabular}} & mclSTExp~\cite{mclstExp}  & 0.1978$_{\pm0.0326}$  & 0.3033$_{\pm0.0216}$  & 0.3098$_{\pm0.1628}$    & 0.0929$_{\pm0.0151}$  & 0.1499$_{\pm0.0814}$  & 0.1065$_{\pm0.0491}$  & 0.1934                     \\
                                                                                  & BLEEP~\cite{bleep}        & 0.1995$_{\pm0.0435}$  & 0.2956$_{\pm0.0253}$  & 0.3221$_{\pm0.1417}$    & 0.0969$_{\pm0.0300}$  & 0.1692$_{\pm0.0729}$  & 0.1336$_{\pm0.0573}$  & 0.2028                     \\ 
\midrule
\multirow{4}{*}{\begin{tabular}[c]{@{}c@{}}$\ours$-\textsc{Adapter}\\(Ours)\end{tabular}}  & \textit{Niche.} + Phikon&0.1925$_{\pm0.0475}$&0.2955$_{\pm0.0347}$&0.4082$_{\pm0.1735}$&0.1912$_{\pm0.0223}$    & 0.2713$_{\pm0.0974}$ & 0.2276$_{\pm0.0644}$ & 0.2644                           \\
                                                                                  & \textit{Niche.} + UNI   & 0.2015$_{\pm0.0461}$   & 0.3097$_{\pm0.0269}$ & \underline{0.4328}$_{\pm0.1621}$    & 0.1903$_{\pm0.0210}$ & 0.2800$_{\pm0.0961}$ & 0.2162$_{\pm0.0600}$ & 0.2718                          \\
                                                                                  & \textit{Visium.} + Phikon &0.2291$_{\pm0.0471}$  & \textbf{0.3368}$_{\pm0.0287}$ & 0.4286$_{\pm0.1758}$    & 0.2133$_{\pm0.0276}$ & \textbf{0.2849}$_{\pm0.0934}$ & 0.2307$_{\pm0.0617}$ & 0.2872                          \\
                                                                                  & \textit{Visium.} + UNI    &\underline{0.2297}$_{\pm0.0466}$  & 0.3318$_{\pm0.0305}$ & 0.4226$_{\pm0.1739}$    & 0.1621$_{\pm0.0290}$ & \underline{0.2848}$_{\pm0.0980}$ & 0.2274$_{\pm0.0635}$ & 0.2764           \\ 
\midrule
\multirow{6}{*}{\begin{tabular}[c]{@{}c@{}}$\ours$-\textsc{Finetune}\\(Ours)\end{tabular}} & \textit{Trans.} + Phikon  &0.2246$_{\pm0.0471}$ &0.3315$_{\pm0.0443}$ &0.4216$_{\pm0.1697}$ &0.2137$_{\pm0.0259}$&0.2389$_{\pm0.0960}$ &0.1726$_{\pm0.0625}$  &0.2672                           \\
                                                                                  & \textit{Trans.} + UNI     &0.1695$_{\pm0.0381}$ & 0.2674$_{\pm0.0236}$ & 0.4276$_{\pm0.1730}$  & 0.1886$_{\pm0.0778}$  & 0.2400$_{\pm0.0897}$ & 0.1726$_{\pm0.0652}$ & 0.2443                          \\
                                                                                  & \textit{Niche.} + Phikon  & 0.2174$_{\pm0.0456}$   & 0.3123$_{\pm0.0278}$ & 0.4194$_{\pm0.1633}$    & 0.2085$_{\pm0.0124}$  & 0.2651$_{\pm0.0973}$ & 0.2155$_{\pm0.0609}$ & 0.2753                     \\
                                                                                  & \textit{Niche.} + UNI     & 0.2045$_{\pm0.0462}$   & 0.3071$_{\pm0.0281}$ & 0.4223$_{\pm0.1599}$    & 0.2102$_{\pm0.0373}$  & 0.2721$_{\pm0.0964}$ & 0.2128$_{\pm0.0641}$ & 0.2715                     \\
                                                                                  & \textit{Visium.} + Phikon & 0.2291$_{\pm0.0516}$   & 0.3291$_{\pm0.0360}$ & \textbf{0.4405}$_{\pm0.1649}$    & \textbf{0.2265}$_{\pm0.0197}$  & 0.2797$_{\pm0.0996}$ & \underline{0.2314}$_{\pm0.0670}$ & \underline{0.2894}                      \\
                                                                                  & \textit{Visium.} + UNI    & \textbf{0.2364}$_{\pm0.0439}$   & \underline{0.3343}$_{\pm0.0363}$ & 0.4317$_{\pm0.1740}$    & \underline{0.2220}$_{\pm0.0211}$  & 0.2843$_{\pm0.1004}$& \textbf{0.2324}$_{\pm0.0689}$ & \textbf{0.2902}                      \\
\bottomrule
\end{tabular}
}
\caption{
\textbf{Results of Gene Expression Prediction. }
The mean and standard deviation of the Pearson correlation coefficient (PCC) for the top 50 highly variable genes (HVG) and highly expressed genes (HEG). \textit{Visium.} refers to Visiumformer, \textit{Niche.} refers to Nicheformer, and \textit{Trans.} indicates a 12-layer Transformer. $\ours$-\textsc{Finetune} and $\ours$-\textsc{Adapter} represent full parameter fine-tuning and the use of adapter.
}
\label{tab:table1}
\end{table*}

\subsection{Downstream Datasets}
Extensive evaluations were conducted across multiple downstream datasets to assess the capabilities of $\ours$ in multimodal and unimodal representation learning. All the downstream evaluation experiments included six datasets and three tasks. These tasks included bimodal gene expression prediction (Section \ref{sec:4.3}), unimodal patch and spot type classification (Section \ref{sec:4.4}), and WSI mutation state prediction (Section \ref{sec:4.5}). 
The six downstream datasets used in these evaluations are as follows: 
\textbf{Human Liver Tissue} (HLT)~\cite{datasetpsc} dataset, comprising four sections and 9,254 paired pathology images and gene expression data.
\textbf{Human Prostate Cancer} (HPC)~\cite{datasethpc} dataset, containing five sections and 14,783 paired samples.
\textbf{HER2-positive breast tumor} (HER2+)\cite{datasether2+} dataset, consisting of 36 sections, with 32 reserved for training, resulting in 11,509 paired samples, as outlined in ST-Net\cite{stnet}.
\textbf{Human Dorsolateral Prefrontal Cortex} (DLPFC)~\cite{datasetdlpfc} dataset, made up of 12 sections and 47,329 paired samples, where each spot was categorized into white matter (WM) and cortical layers L1–L6.
\textbf{Human Breast Cancer}~\cite{datasetbreast} (10X Breast) dataset, with one section and 3,789 paired samples, where each spot was categorized into four tissue subtypes.
\textbf{LUAD-mutation} dataset, which includes 692 Fresh Frozen WSIs from 437 patients in TCGA-LUAD, used to predict mutation status (positive/negative) for four specific genes: EGFR, KRAS, STK11, and TP53, as detailed in DeepPATH~\cite{coudray2018classification}. 
For details on the downstream datasets, comparison methods, and downstream model training, please refer to \textbf{Appendix} \ref{B}.

\subsection{Multimodal Representation Learning}
\label{sec:4.3}
The multimodal representation capability of $\ours$ is evaluated through a bimodal gene expression prediction task (Figure \ref{fig:fig2}\textcolor{red}{a}).  As shown in Table \ref{tab:table1}, $\ours$ was assessed on three datasets: Human Liver Tissue dataset (HLT), Human Prostate Cancer dataset (HPC), and HER2-positive breast tumor dataset (HER2+). HLT and HPC were measured using the Visium platform~\cite{visium}. The HER2+ dataset, derived from the Spatial Transcriptomics platform~\cite{staahl2016visualization}, was then used to assess the transfer learning capabilities of $\ours$ across different technologies and platforms. 

Specifically, this task aims to predict full-dimensional gene expression based on pathology images. Two strategies were employed for evaluation: $\ours$-\textsc{Finetune} (full-parameter fine-tuning) and $\ours$-\textsc{Adapter}, which adds two trainable linear layers with ReLU activation to both the frozen encoders. Additionally, we included other task-specific methods, including regression-based and contrastive learning-based models. The average Pearson correlation coefficient (PCC)~\cite{cohen2009pearson} was reported for the top 50 highly variable genes (HVG) and highly expressed genes (HEG), utilizing a leave-one-out cross-validation method. To eliminate data leakage and ensure a fair comparison, the datasets used in this section were not included in the pre-training phase.
\begin{figure*}[tp]
    \centering
    \includegraphics[width=1\linewidth]{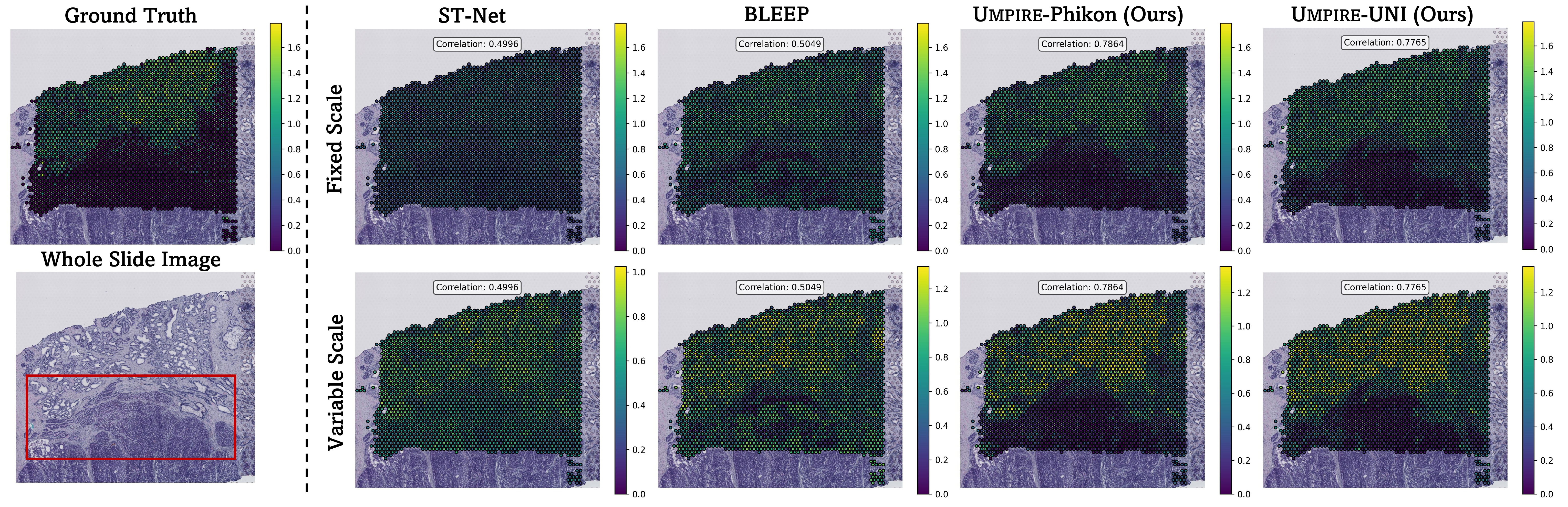}
    \caption{
    \textbf{Visualization of Bimodal Gene Expression Prediction. } 
    Ground truth and predicted spatially resolved expression levels for PIBF1 overlaying the whole slide image of sample patient-1-H2-5, visualized with a fixed (top) and a variable (bottom) color scale.}
    \label{fig:fig3}
\end{figure*}

\vspace{0.5pt}\noindent\textbf{Regression-based \emph{vs.} Contrastive Learning-based:} When predicting full-dimensional gene expression, regression-based methods often face the ``curse of dimensionality", causing training failures for all but ST-Net. In contrast, contrastive learning methods, using the Query-Reference paradigm, excel in full-dimensional prediction. The top-performing contrastive learning method, BLEEP, shows an average improvement of $+60.1\%$ and $+46.4\%$ over ST-Net on the HPC and HER2+ datasets, respectively.

\vspace{0.5pt}\noindent\textbf{$\ours$ \emph{vs.} Contrastive Learning-based:} Compared to contrastive learning-based methods, both $\ours$-\textsc{Adapter} and $\ours$-\textsc{Finetune} demonstrate significant improvements. Specifically, $\ours$-\textsc{Adapter} achieves an average increase of $+39.0\%$ over BLEEP, while $\ours$-\textsc{Finetune} shows an improvement of $+42.9\%$. Apart from the HLT and HPC datasets based on Visium, $\ours$ also achieved outstanding performance on the HER2+ dataset, with an average improvement of $+83.8\%$. The HER2+ dataset was sequenced using the Spatial Transcriptomics platform, which was not encountered during the pre-training phase. This reflects the strong generalization capabilities of $\ours$, which benefit from the diversity of data used during pre-training. $\ours$ performs well across various organs (liver, prostate, and breast), disease states (healthy and cancerous), and sequencing platforms (Visium and Spatial Transcriptomics). To further demonstrate that the significant performance improvement of $\ours$ is not solely attributable to a more powerful vision encoder, we replaced the vision encoders of ST-Net and BLEEP with Phikon. This modification leads to an improvement in the performance of ST-Net; however, it still significantly lags behind the original BLEEP. Applying the same operation to BLEEP results in a performance decrease of about $-58.2\%$ due to the small training dataset, which is unsuitable for large-parameter vision encoders (please refer to \textbf{Appendix} \ref{A.5}).

\vspace{0.5pt}\noindent\textbf{\textit{Visium.} \emph{vs.} \textit{Niche.} \emph{vs.} \textit{Trans.}:} For comparison, we employed three different gene encoders: our Visiumformer (\textit{Visium.}), Nicheformer (\textit{Niche.}) pre-trained on 100M single-cell and spatial transcriptomics data, and a randomly initialized 12-layer Transformer (\textit{Trans.}). The \textit{Trans.} encoder utilizes continuous gene expression values from the top 1500 highly variable genes, while both \textit{Visium.} and \textit{Niche.} require tokenization. Notably, our \textit{Visium.} combined with vision encoders consistently outperforms the others, achieving an average PCC that is \( +6\% \) higher than that of \textit{Niche.} and \( +13.3\% \) higher than that of \textit{Trans.}. Although \textit{Niche.} is not pre-trained on Visium data, it performs well after multimodal alignment pre-training. In contrast, \textit{Trans.} underperforms due to the lack of pre-training in the first phase, despite participating in the second pre-training phase.

\vspace{0.5pt}\noindent\textbf{$\ours$-\textsc{Adapter} \emph{vs.} $\ours$-\textsc{Finetune}:} A key advantage of pre-trained models is their efficient performance with minimal resources, achieved through small-parameter fine-tuning on downstream tasks~\cite{clip,shi2024vila}. To leverage this capability, $\ours$-\textsc{Adapter} was introduced. Overall, the $\ours$-\textsc{Adapter} performs worse than $\ours$-\textsc{Finetune} by an average of $-2.8\%$. However, the $\ours$-\textsc{Adapter} uses only $0.3\%$ to $0.8\%$ of the trainable parameters required by $\ours$-\textsc{Finetune}. For individual datasets, the $\ours$-\textsc{Adapter} lags behind $\ours$-\textsc{Finetune} by $-7.1\%$ on the larger HPC dataset, while it nearly matches $\ours$-\textsc{Finetune} on the smaller HLT and HER2+ datasets. We recommend the $\ours$-\textsc{Adapter} for limited data or computational resources and $\ours$-\textsc{Finetune} for other scenarios to leverage $\ours$ fully.

\vspace{0.5pt}\noindent\textbf{Case Study:} We visualized the actual expression of PIBF1 (Figure \ref{fig:fig3}) and CTSC (see \textbf{Appendix} \ref{A.2}) in the sample HPC-patient-1-H2-5, along with the expression predicted by various methods. PIBF1 and CTSC are known to influence cell proliferation and autophagy, each playing distinct roles in tumor invasion and metastasis~\cite{li2024pibf1,xiao2021cathepsin}. $\ours$ shows greater biological heterogeneity within the slices compared to ST-Net and BLEEP, especially between the tumor and the normal tissue (see red box in Figure \ref{fig:fig3}).

\begin{figure*}[tp]
    \centering
    \includegraphics[width=1\linewidth]{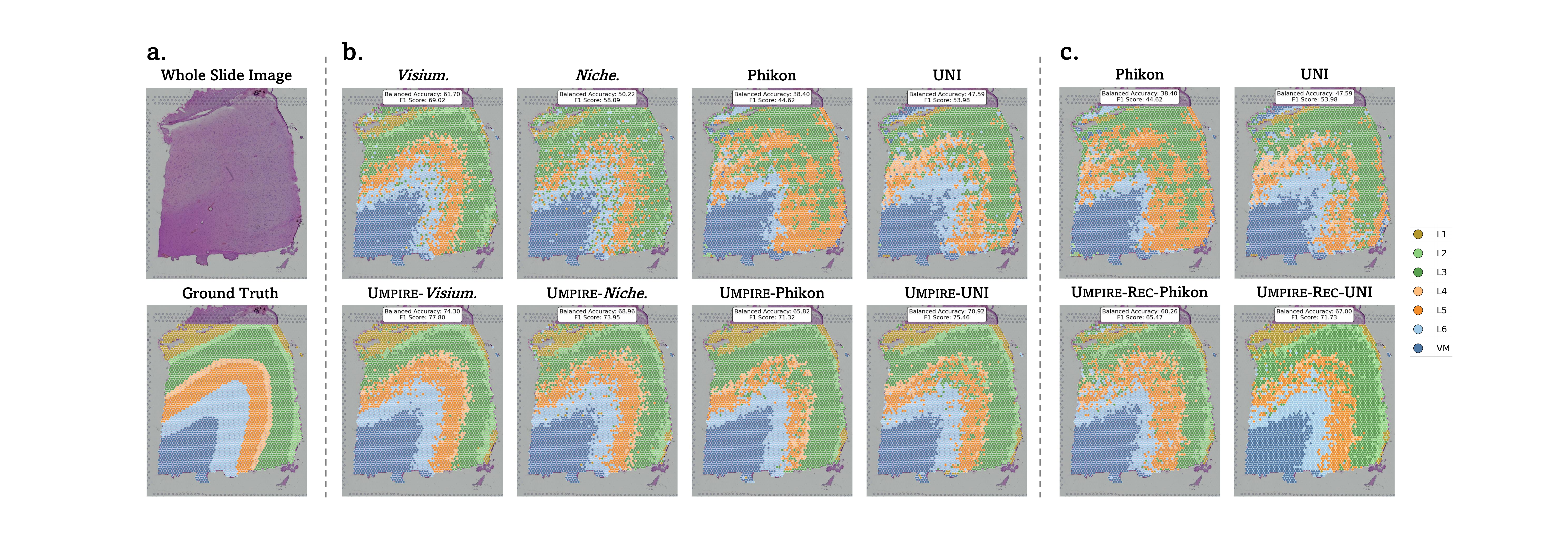}
    \caption{
    \textbf{Visualization of Linear Probing. } 
    \textbf{a.} Whole Slide Image and Ground Truth; 
    \textbf{b.} Predicted spot/patch types for sample 151673, visualized before (top) and after (bottom) multimodal pre-training with contrastive loss; 
    \textbf{c.} with reconstruction loss.
    }
    \label{fig:fig4}
\end{figure*}

\begin{table}
\centering
\resizebox{0.48\textwidth}{!}{
\begin{tabular}{l|c|cccc} 
\toprule
\multicolumn{1}{c|}{\multirow{2}{*}{Method}} & \multirow{2}{*}{Modality} & \multicolumn{2}{c}{\textbf{DLPFC}}                & \multicolumn{2}{c}{\textbf{10X Breast}}  \\
\multicolumn{1}{c|}{}                        &                              & Bal. Acc.                 & \multicolumn{1}{c|}{Wgt. F1}  & Bal. Acc. & Wgt. F1             \\ 
\midrule
GeneMLP                                      & \multirow{3}{*}{$\mathcal{G}$}& 53.46$_{\pm1.77}$        & 64.13$_{\pm2.83}$             & 75.95$_{\pm1.90}$         & 76.27$_{\pm1.59}$               \\
\textit{Niche.}~\cite{schaar2024nicheformer} &                              & 45.12$_{\pm4.50}$         & 56.18$_{\pm3.44}$             & 72.56$_{\pm1.49}$         & 74.80$_{\pm1.51}$               \\
\textit{Visium.}(Ours)                       &                              & 55.13$_{\pm4.11}$         & 65.87$_{\pm3.86}$             & 76.97$_{\pm1.95}$         & 77.54$_{\pm1.66}$               \\ 
\midrule
Phikon~\cite{filiot2023phikon}               & \multirow{2}{*}{$\mathcal{P}$}& 48.17$_{\pm10.76}$       & 56.92$_{\pm8.89}$             & 82.10$_{\pm2.35}$         & 83.04$_{\pm2.41}$               \\
UNI~\cite{chen2024uni}                       &                              & 53.72$_{\pm10.59}$        & 62.84$_{\pm7.12}$             & 81.88$_{\pm4.08}$         & 82.92$_{\pm3.83}$               \\ 
\midrule
$\ours$-\textsc{Rec}-Phikon                          &\multirow{6}{*}{$\mathcal{G+P}$}& 54.00$_{\pm7.70}$       & 64.10$_{\pm4.21}$             & 75.48$_{\pm3.73}$         & 75.23$_{\pm2.98}$                     \\
$\ours$-\textsc{Rec}-UNI                             &                              & 60.59$_{\pm 8.27}$        & 69.88$_{\pm 4.07}$            & 76.03$_{\pm2.35}$         & 77.71$_{\pm1.69}$                     \\
$\ours$-Phikon                               &                              & 68.53$_{\pm7.14}$         & 76.34$_{\pm4.19}$             & \textbf{85.06}$_{\pm1.19}$ & \textbf{86.07}$_{\pm1.17}$               \\
$\ours$-UNI                                  &                    & \underline{68.76}$_{\pm8.17}$       & \underline{76.83}$_{\pm3.89}$ & \underline{84.31}$_{\pm2.98}$ & \underline{85.51}$_{\pm2.48}$               \\
$\ours$-\textit{Niche.}                      &                              & 68.69$_{\pm3.87}$         & 76.59$_{\pm3.17}$             & 79.20$_{\pm2.37}$         & 80.39$_{\pm1.47}$                    \\
$\ours$-\textit{Visium.}                     &                             & \textbf{70.70}$_{\pm3.21}$ & \textbf{77.97}$_{\pm2.76}$    & 82.06$_{\pm1.45}$         & 83.17$_{\pm1.23}$                    \\
\bottomrule
\end{tabular}
}
\caption{
\textbf{Results of Linear Probing. }
The average and standard deviation (in $\%$) of balanced accuracy (Bal. Acc.) and F1 score (Wgt. F1) are reported for DLPFC and 10X Breast. 
$\mathcal{G}$ indicates pre-training on gene data, $\mathcal{P}$ indicates pre-training on pathological images, and $\mathcal{G+P}$ signifies multimodal joint pre-training.}
\label{tab:table2}
\end{table}

\subsection{Linear Probing Classification}
\label{sec:4.4}
Table \ref{tab:table2} presents the evaluation results of $\ours$ for classifying human dorsolateral prefrontal cortex (DLPFC) morphotypes and human breast cancer (10X Breast). Different brain regions show subtle visual differences, so gene expression data is typically used for spot classification. We use DLPFC to evaluate how well $\ours$ integrates complementary information across modalities. In contrast, the 10X Breast dataset exhibits significant visual differences between tissue types, allowing effective classification using visual information alone. This dataset helps assess whether the molecular perspective introduced by $\ours$ harms the original vision encoder. Following standard practices in SSL~\cite{oquab2023dinov2,darcet2023vitneedreg,caron2021emerging}, linear probing was employed to benchmark $\ours$, $\ours$-$\textsc{Rec}$, and Visiumformer (Figure \ref{fig:fig2}\textcolor{red}{b}). We also benchmarked Nicheformer~\cite{schaar2024nicheformer}, Phikon~\cite{filiot2023phikon} and UNI~\cite{chen2024uni} for comparison.

\vspace{0.5pt}\noindent\textbf{Gene-based \emph{vs.} Image-based \emph{vs.} $\ours$-based:} We evaluated three categories of models: gene-based ($\mathcal{G}$), pathology image-based ($\mathcal{P}$), and multimodal pre-trained models ($\mathcal{G+P}$). Gene-based models perform well on DLPFC; however, their performance declines in dataset with significant visual variations, \ie 10X Breast. Regardless of the modality utilized, pre-training with $\ours$ consistently enhances model performance. Following alignment, Visiumformer demonstrated a balanced accuracy increase of $+28.2\%$ on DLPFC and $+6.6\%$ on 10X Breast. For the vision encoders Phikon and UNI, balanced accuracy improved by up to $+42.3\%$ on DLPFC and $+3.6\%$ on 10X Breast. These experiments clearly demonstrate that $\ours$ benefits from multimodal alignment, effectively enhancing information complementarity and significantly boosting performance.

\vspace{0.5pt}\noindent\textbf{$\ours$ \emph{vs.} $\ours$-$\textsc{Rec}$:}  In contrast to the improved consistency of $\ours$ across all datasets, the vision encoders pre-trained with reconstruction loss ($\ours$-$\textsc{Rec}$) demonstrated significant performance improvements on DLPFC but experienced varying degrees of decline on 10X Breast. We believe that the high sparsity and dimensionality of gene expression restrict the vision encoders' ability to learn effectively from the gene modality when utilizing regression and reconstruction methods.

\vspace{0.5pt}\noindent\textbf{\textit{Visium.} \emph{vs.} \textit{Niche.} \emph{vs.} GeneMLP:}  Analogous to the \textit{Trans.} described in Section \ref{sec:4.3}, an unpretrained GeneMLP was established as a baseline. In accordance with standard linear probing protocols, GeneMLP selects the top 1,500 highly variable gene expressions after normalization, which are subsequently processed through a linear layer for classification. Pre-training on ViSTomics-4M enabled Visiumformer to outperform other gene-based models. Conversely, Nicheformer, which lacked access to Visium platform gene expression during pre-training, performed worse than GeneMLP. After alignment, both models exhibited noticeable improvements; however, $\ours$-\textit{Niche.} still fell short of $\ours$-\textit{Visium.}. This underscores that while alignment can enhance performance, it cannot fully compensate for the degradation caused by the absence of corresponding data in the initial stage. This necessity prompted the development of ViSTomics-4M and the pre-training of Visiumformer.

\begin{figure}[tp]
    \centering
    \includegraphics[width=\linewidth]{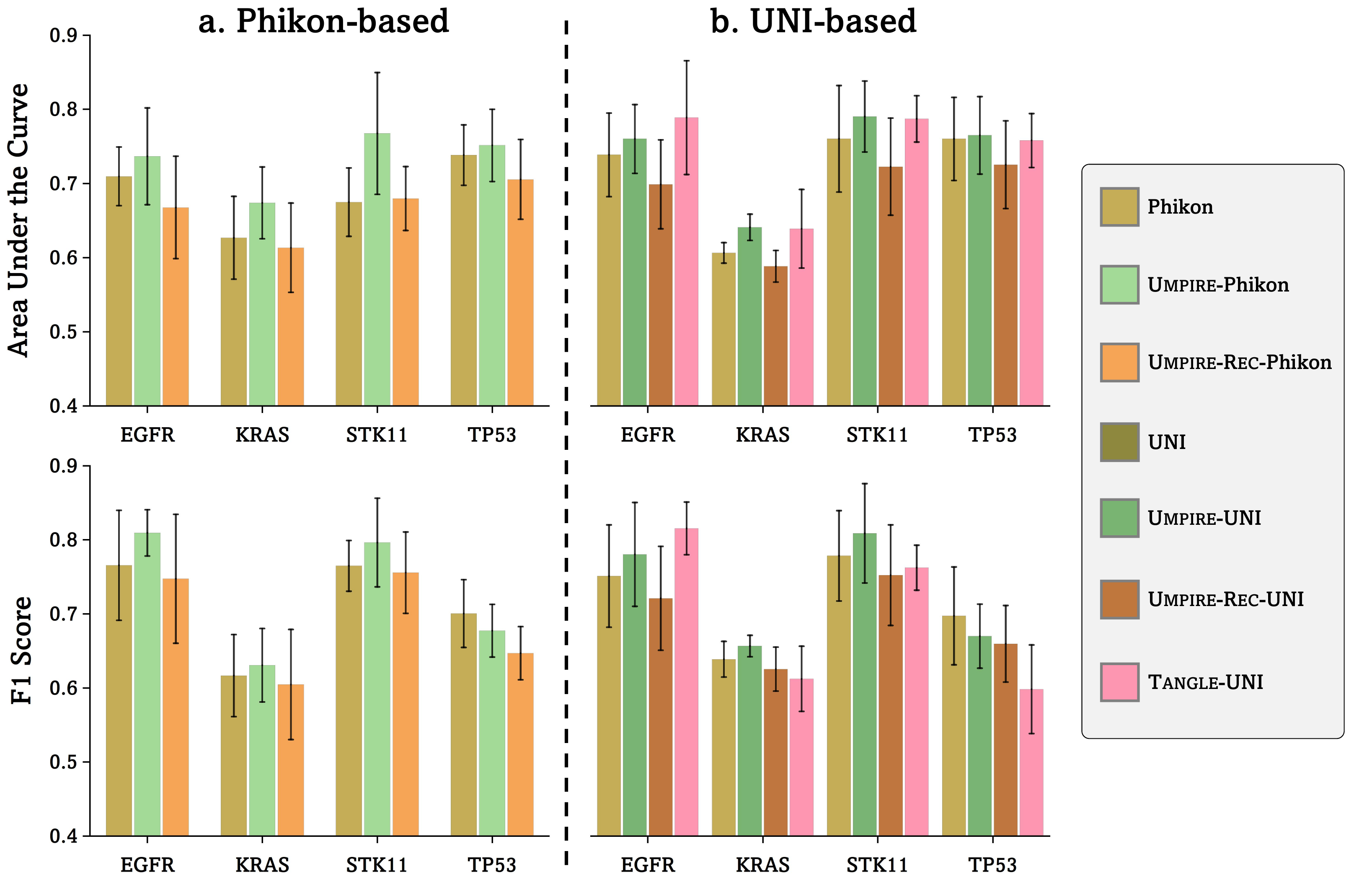}
    \caption{\textbf{MIL-based WSI Classification}. Comparison of $\ours$ and baselines for WSI-level gene mutation state classification using MIL. \textbf{a.} Based on Phikon. \textbf{b.} Based on UNI.}
    \label{fig:fig5}
\end{figure}

\vspace{0.5pt}\noindent\textbf{Case Study:} Figure \ref{fig:fig4} visualizes the linear probing classification results for sample 151673 from DLPFC, illustrating performance before (top) and after (bottom) multimodal pre-training. After pre-training, the model's ability to differentiate between different brain regions significantly improves across both modalities, particularly among layers L1 to L6. This demonstrates that our $\ours$ effectively integrates information from both modalities, achieving a synergistic effect in which the combined performance exceeds the sum of the individual contributions.

\vspace{0.5pt}\noindent\textbf{Zero-shot Embedding:} Following multimodal pre-training, we performed zero-shot embedding visualization to analyze the embeddings of the two modalities before and after pre-training using t-Distributed Stochastic Neighbor Embedding (t-SNE)~\cite{van2008visualizingtsne}. DLPFC served as a benchmark for computing two clustering quality metrics, including the Silhouette score~\cite{rousseeuw1987silhouettes} and the Davies-Bouldin index~\cite{davies1979dbindex}. The t-SNE visualizations and corresponding evaluation metrics are provided in \textbf{Appendix} \ref{A.3}. Our results indicate that the embeddings after pre-training ($\ours$ and $\ours$-$\textsc{Rec}$) are more effective at distinguishing various brain regions. Notably, pre-training enhances model performance on pathology images while also improving the results on gene expression. The integration of gene expression with pathology images further enhances the model's ability to discern subtle features within the images and reveals previously unrecognized insights from the gene expression.

\vspace{0.5pt}\noindent\textbf{Loss Ablation:} Ablation studies were conducted on DLPFC to evaluate the impact of different loss functions. When the symmetric contrastive loss was replaced with regression-based loss functions (mean squared error and L1 loss), the weighted F1 score for Phikon decreased by $-16.0\%$ and $-16.4\%$, respectively. We reasonably attribute this decline to the high sparsity of the gene expression, which negatively impacts reconstruction performance. Additionally, the effects of replacing the symmetric contrastive loss with either unilateral contrastive loss~\cite{clip} or InfoNCE loss~\cite{he2020momentum} were investigated, both of which resulted in varying degrees of performance degradation (see \textbf{Appendix} \ref{A.4}).

\subsection{MIL-based WSI Classification}
\label{sec:4.5}
Certain cancer analyses require global WSI information; however, the large size of WSIs necessitates using Multiple Instance Learning (MIL) for WSI-level tasks. The impact of $\ours$ on WSI-level performance across four WSI gene mutation status classification tasks was evaluated. All tasks utilized ABMIL~\cite{ilse2018attention} as the instance aggregation method (see Figure \ref{fig:fig2}\textcolor{red}{c}). All experiments were conducted using five-fold cross-validation at the patient level.

Figure \ref{fig:fig5} compares the performance of Phikon and UNI before and after alignment. Despite being self-supervised on numerous WSIs, Phikon and UNI exhibit suboptimal results in this challenging task. $\ours$ outperformed the original vision encoder in three sub-tasks, achieving maximum relative improvements of $+13.7\%$ in AUC and $+7.7\%$ in the F1 Score. In contrast, $\ours$-$\textsc{Rec}$ significantly underperformed compared to the original encoder. We speculate that the regression-based pre-training method caused the vision encoder to focus excessively on gene-level features, diminishing its ability to capture the original semantic information from the images. Conversely, our $\ours$ employs a contrastive learning approach that enhances the vision encoder's ability to capture gene-level details while preserving its capacity to retain visual semantic information.

TANGLE~\cite{tangle} focuses on pre-training at the WSI level, using UNI~\cite{chen2024uni} as a feature extractor and ABMIL as an aggregation module to align WSIs with bulk RNA data across 27 TCGA cohorts. To adapt TANGLE for WSI classification tasks, we utilize a frozen UNI to extract features and apply the pre-trained aggregation module from TANGLE. Our experimental results show that $\ours$ outperforms TANGLE in three out of four sub-tasks (see Figure \ref{fig:fig5}\textcolor{red}{b}). In the sub-tasks involving KRAS, STK11, and TP53, $\ours$ demonstrates comparable performance to TANGLE in terms of AUC, surpassing it by an average of $+0.55\%$, while achieving an average improvement of $+8.44\%$ in F1 Score. In the EGFR sub-task, $\ours$ falls short, lagging behind TANGLE by $-3.65\%$ and $-4.30\%$, respectively.

\section{Conclusion and Discussion}
\label{sec:conclusion}
\vspace{0.5pt}\noindent\textbf{Conclusion:}
In this paper, we first collected and constructed the largest Visium-based spatial transcriptomics (ST) dataset and then introduced a unified molecule-enhanced pathology image representation learning framework. Our approach, $\ours$, employs a two-stage pre-training process on extensive ST data and paired pathology image-ST gene expression. Comprehensive evaluations of $\ours$ were conducted across multiple downstream tasks, demonstrating its significant superiority over various baseline methods in all tasks. As the first attempt at a molecule-enhanced pathology image representation learning framework, $\ours$ will also serve as a foundational model for future research.

\vspace{0.5pt}\noindent\textbf{Future Work:}
These results underscore the potential of multimodal pre-training, paving the way for future advancements. Compared to other visual-language pre-training methods, the data used remains relatively small~\cite{conch,plip}, and future work should focus on larger-scale data collection. Additionally, while we demonstrated that models pre-trained on Visium data can be effectively transferred to other sequencing platforms, subsequent research should aim to develop a more generalized and robust model encompassing multiple sequencing technologies and platforms~\cite{schaar2024nicheformer}.

\section{Acknowledgment}

This project was funded by the National Natural Science Foundation of China 82090052.


{\small
\bibliographystyle{ieeenat_fullname}
\bibliography{_references}

\begin{thebibliography}{85}
\providecommand{\natexlab}[1]{#1}
\providecommand{\url}[1]{\texttt{#1}}
\expandafter\ifx\csname urlstyle\endcsname\relax
  \providecommand{\doi}[1]{doi: #1}\else
  \providecommand{\doi}{doi: \begingroup \urlstyle{rm}\Url}\fi

\bibitem[Andersson et~al.(2021)Andersson, Larsson, Stenbeck, Salm{\'e}n, Ehinger, Wu, Al-Eryani, Roden, Swarbrick, Borg, et~al.]{datasether2+}
Alma Andersson, Ludvig Larsson, Linnea Stenbeck, Fredrik Salm{\'e}n, Anna Ehinger, Sunny~Z Wu, Ghamdan Al-Eryani, Daniel Roden, Alex Swarbrick, {\AA}ke Borg, et~al.
\newblock Spatial deconvolution of her2-positive breast cancer delineates tumor-associated cell type interactions.
\newblock \emph{Nature communications}, 12\penalty0 (1):\penalty0 6012, 2021.

\bibitem[Andrews et~al.(2024)Andrews, Nakib, Perciani, Ma, Liu, Winter, Camat, Chung, Lumanto, Manuel, et~al.]{datasetpsc}
Tallulah~S Andrews, Diana Nakib, Catia~T Perciani, Xue~Zhong Ma, Lewis Liu, Erin Winter, Damra Camat, Sai~W Chung, Patricia Lumanto, Justin Manuel, et~al.
\newblock Single-cell, single-nucleus, and spatial transcriptomics characterization of the immunological landscape in the healthy and psc human liver.
\newblock \emph{Journal of Hepatology}, 80\penalty0 (5):\penalty0 730--743, 2024.

\bibitem[Azizi et~al.(2023)Azizi, Culp, Freyberg, Mustafa, Baur, Kornblith, Chen, Tomasev, Mitrovi{\'c}, Strachan, et~al.]{azizi2023robust}
Shekoofeh Azizi, Laura Culp, Jan Freyberg, Basil Mustafa, Sebastien Baur, Simon Kornblith, Ting Chen, Nenad Tomasev, Jovana Mitrovi{\'c}, Patricia Strachan, et~al.
\newblock Robust and data-efficient generalization of self-supervised machine learning for diagnostic imaging.
\newblock \emph{Nature Biomedical Engineering}, 7\penalty0 (6):\penalty0 756--779, 2023.

\bibitem[Brown(2020)]{gpt3}
Tom~B Brown.
\newblock Language models are few-shot learners.
\newblock \emph{arXiv preprint arXiv:2005.14165}, 2020.

\bibitem[Caron et~al.(2021)Caron, Touvron, Misra, J{\'e}gou, Mairal, Bojanowski, and Joulin]{caron2021emerging}
Mathilde Caron, Hugo Touvron, Ishan Misra, Herv{\'e} J{\'e}gou, Julien Mairal, Piotr Bojanowski, and Armand Joulin.
\newblock Emerging properties in self-supervised vision transformers.
\newblock In \emph{Proceedings of the IEEE/CVF international conference on computer vision}, pages 9650--9660, 2021.

\bibitem[Chen et~al.(2024{\natexlab{a}})Chen, Zhou, Wu, Zhang, Li, and Li]{chen2024stimage}
Jiawen Chen, Muqing Zhou, Wenrong Wu, Jinwei Zhang, Yun Li, and Didong Li.
\newblock Stimage-1k4m: A histopathology image-gene expression dataset for spatial transcriptomics.
\newblock \emph{arXiv preprint arXiv:2406.06393}, 2024{\natexlab{a}}.

\bibitem[Chen et~al.(2015)Chen, Boettiger, Moffitt, Wang, and Zhuang]{merfish}
Kok~Hao Chen, Alistair~N Boettiger, Jeffrey~R Moffitt, Siyuan Wang, and Xiaowei Zhuang.
\newblock Spatially resolved, highly multiplexed rna profiling in single cells.
\newblock \emph{Science}, 348\penalty0 (6233):\penalty0 aaa6090, 2015.

\bibitem[Chen et~al.(2024{\natexlab{b}})Chen, Ding, Lu, Williamson, Jaume, Song, Chen, Zhang, Shao, Shaban, et~al.]{chen2024uni}
Richard~J Chen, Tong Ding, Ming~Y Lu, Drew~FK Williamson, Guillaume Jaume, Andrew~H Song, Bowen Chen, Andrew Zhang, Daniel Shao, Muhammad Shaban, et~al.
\newblock Towards a general-purpose foundation model for computational pathology.
\newblock \emph{Nature Medicine}, 30\penalty0 (3):\penalty0 850--862, 2024{\natexlab{b}}.

\bibitem[Chen et~al.(2020)Chen, Kornblith, Norouzi, and Hinton]{simclr}
Ting Chen, Simon Kornblith, Mohammad Norouzi, and Geoffrey Hinton.
\newblock A simple framework for contrastive learning of visual representations.
\newblock In \emph{International conference on machine learning}, pages 1597--1607. PMLR, 2020.

\bibitem[Chen et~al.(2021)Chen, Xie, and He]{mocov3}
Xinlei Chen, Saining Xie, and Kaiming He.
\newblock An empirical study of training self-supervised vision transformers.
\newblock In \emph{Proceedings of the IEEE/CVF international conference on computer vision}, pages 9640--9649, 2021.

\bibitem[Cohen et~al.(2009)Cohen, Huang, Chen, Benesty, Benesty, Chen, Huang, and Cohen]{cohen2009pearson}
Israel Cohen, Yiteng Huang, Jingdong Chen, Jacob Benesty, Jacob Benesty, Jingdong Chen, Yiteng Huang, and Israel Cohen.
\newblock Pearson correlation coefficient.
\newblock \emph{Noise reduction in speech processing}, pages 1--4, 2009.

\bibitem[Coudray et~al.(2018)Coudray, Ocampo, Sakellaropoulos, Narula, Snuderl, Feny{\"o}, Moreira, Razavian, and Tsirigos]{coudray2018classification}
Nicolas Coudray, Paolo~Santiago Ocampo, Theodore Sakellaropoulos, Navneet Narula, Matija Snuderl, David Feny{\"o}, Andre~L Moreira, Narges Razavian, and Aristotelis Tsirigos.
\newblock Classification and mutation prediction from non--small cell lung cancer histopathology images using deep learning.
\newblock \emph{Nature medicine}, 24\penalty0 (10):\penalty0 1559--1567, 2018.

\bibitem[Cui et~al.(2024)Cui, Wang, Maan, Pang, Luo, Duan, and Wang]{cui2024scgpt}
Haotian Cui, Chloe Wang, Hassaan Maan, Kuan Pang, Fengning Luo, Nan Duan, and Bo Wang.
\newblock scgpt: toward building a foundation model for single-cell multi-omics using generative ai.
\newblock \emph{Nature Methods}, pages 1--11, 2024.

\bibitem[Darcet et~al.(2023)Darcet, Oquab, Mairal, and Bojanowski]{darcet2023vitneedreg}
Timothée Darcet, Maxime Oquab, Julien Mairal, and Piotr Bojanowski.
\newblock Vision transformers need registers, 2023.

\bibitem[Davies and Bouldin(1979)]{davies1979dbindex}
David~L Davies and Donald~W Bouldin.
\newblock A cluster separation measure.
\newblock \emph{IEEE transactions on pattern analysis and machine intelligence}, \penalty0 (2):\penalty0 224--227, 1979.

\bibitem[Devlin(2018)]{bert}
Jacob Devlin.
\newblock Bert: Pre-training of deep bidirectional transformers for language understanding.
\newblock \emph{arXiv preprint arXiv:1810.04805}, 2018.

\bibitem[Ding et~al.(2023)Ding, Zhou, Metaxas, and Zhang]{ding2023pathology}
Kexin Ding, Mu Zhou, Dimitris~N Metaxas, and Shaoting Zhang.
\newblock Pathology-and-genomics multimodal transformer for survival outcome prediction.
\newblock In \emph{MICCAI}, pages 622--631. Springer, 2023.

\bibitem[Elhanani et~al.(2023)Elhanani, Ben-Uri, and Keren]{elhanani2023spatial}
Ofer Elhanani, Raz Ben-Uri, and Leeat Keren.
\newblock Spatial profiling technologies illuminate the tumor microenvironment.
\newblock \emph{Cancer cell}, 41\penalty0 (3):\penalty0 404--420, 2023.

\bibitem[Erickson et~al.(2022)Erickson, He, Berglund, Marklund, Mirzazadeh, Schultz, Kvastad, Andersson, Bergenstr{\aa}hle, Bergenstr{\aa}hle, et~al.]{datasethpc}
Andrew Erickson, Mengxiao He, Emelie Berglund, Maja Marklund, Reza Mirzazadeh, Niklas Schultz, Linda Kvastad, Alma Andersson, Ludvig Bergenstr{\aa}hle, Joseph Bergenstr{\aa}hle, et~al.
\newblock Spatially resolved clonal copy number alterations in benign and malignant tissue.
\newblock \emph{Nature}, 608\penalty0 (7922):\penalty0 360--367, 2022.

\bibitem[Filiot et~al.(2023)Filiot, Ghermi, Olivier, Jacob, Fidon, Mac~Kain, Saillard, and Schiratti]{filiot2023phikon}
Alexandre Filiot, Ridouane Ghermi, Antoine Olivier, Paul Jacob, Lucas Fidon, Alice Mac~Kain, Charlie Saillard, and Jean-Baptiste Schiratti.
\newblock Scaling self-supervised learning for histopathology with masked image modeling.
\newblock \emph{medRxiv}, pages 2023--07, 2023.

\bibitem[Glaser et~al.(2017)Glaser, Reder, Chen, McCarty, Yin, Wei, Wang, True, and Liu]{glaser2017light}
Adam~K Glaser, Nicholas~P Reder, Ye Chen, Erin~F McCarty, Chengbo Yin, Linpeng Wei, Yu Wang, Lawrence~D True, and Jonathan~TC Liu.
\newblock Light-sheet microscopy for slide-free non-destructive pathology of large clinical specimens.
\newblock \emph{Nature biomedical engineering}, 1\penalty0 (7):\penalty0 0084, 2017.

\bibitem[Gr{\"u}n(2020)]{grun2020revealing}
Dominic Gr{\"u}n.
\newblock Revealing dynamics of gene expression variability in cell state space.
\newblock \emph{Nature methods}, 17\penalty0 (1):\penalty0 45--49, 2020.

\bibitem[Gu et~al.(2023)Gu, Dai, Lu, and Zhao]{gu2023comprehensive}
Jianlei Gu, Jiawei Dai, Hui Lu, and Hongyu Zhao.
\newblock Comprehensive analysis of ubiquitously expressed genes in humans from a data-driven perspective.
\newblock \emph{Genomics, Proteomics \& Bioinformatics}, 21\penalty0 (1):\penalty0 164--176, 2023.

\bibitem[Hadsell et~al.(2006)Hadsell, Chopra, and LeCun]{hadsell2006dimensionality}
Raia Hadsell, Sumit Chopra, and Yann LeCun.
\newblock Dimensionality reduction by learning an invariant mapping.
\newblock In \emph{2006 IEEE computer society conference on computer vision and pattern recognition (CVPR'06)}, pages 1735--1742. IEEE, 2006.

\bibitem[He et~al.(2020{\natexlab{a}})He, Bergenstr{\aa}hle, Stenbeck, Abid, Andersson, Borg, Maaskola, Lundeberg, and Zou]{stnet}
Bryan He, Ludvig Bergenstr{\aa}hle, Linnea Stenbeck, Abubakar Abid, Alma Andersson, {\AA}ke Borg, Jonas Maaskola, Joakim Lundeberg, and James Zou.
\newblock Integrating spatial gene expression and breast tumour morphology via deep learning.
\newblock \emph{Nature biomedical engineering}, 4\penalty0 (8):\penalty0 827--834, 2020{\natexlab{a}}.

\bibitem[He et~al.(2016)He, Zhang, Ren, and Sun]{resnet}
Kaiming He, Xiangyu Zhang, Shaoqing Ren, and Jian Sun.
\newblock Deep residual learning for image recognition.
\newblock In \emph{Proceedings of the IEEE conference on computer vision and pattern recognition}, pages 770--778, 2016.

\bibitem[He et~al.(2020{\natexlab{b}})He, Fan, Wu, Xie, and Girshick]{he2020momentum}
Kaiming He, Haoqi Fan, Yuxin Wu, Saining Xie, and Ross Girshick.
\newblock Momentum contrast for unsupervised visual representation learning.
\newblock In \emph{Proceedings of the IEEE/CVF conference on computer vision and pattern recognition}, pages 9729--9738, 2020{\natexlab{b}}.

\bibitem[He et~al.(2022)He, Chen, Xie, Li, Doll{\'a}r, and Girshick]{he2022masked}
Kaiming He, Xinlei Chen, Saining Xie, Yanghao Li, Piotr Doll{\'a}r, and Ross Girshick.
\newblock Masked autoencoders are scalable vision learners.
\newblock In \emph{Proceedings of the IEEE/CVF conference on computer vision and pattern recognition}, pages 16000--16009, 2022.

\bibitem[Huang et~al.(2017)Huang, Liu, Van Der~Maaten, and Weinberger]{DenseNet}
Gao Huang, Zhuang Liu, Laurens Van Der~Maaten, and Kilian~Q Weinberger.
\newblock Densely connected convolutional networks.
\newblock In \emph{Proceedings of the IEEE conference on computer vision and pattern recognition}, pages 4700--4708, 2017.

\bibitem[Huang et~al.(2023)Huang, Bianchi, Yuksekgonul, Montine, and Zou]{plip}
Zhi Huang, Federico Bianchi, Mert Yuksekgonul, Thomas~J Montine, and James Zou.
\newblock A visual--language foundation model for pathology image analysis using medical twitter.
\newblock \emph{Nature Medicine}, pages 1--10, 2023.

\bibitem[Ilse et~al.(2018)Ilse, Tomczak, and Welling]{ilse2018attention}
Maximilian Ilse, Jakub Tomczak, and Max Welling.
\newblock Attention-based deep multiple instance learning.
\newblock In \emph{International conference on machine learning}, pages 2127--2136. PMLR, 2018.

\bibitem[Jain and Eadon(2024)]{jain2024spatial}
Sanjay Jain and Michael~T Eadon.
\newblock Spatial transcriptomics in health and disease.
\newblock \emph{Nature Reviews Nephrology}, pages 1--13, 2024.

\bibitem[Janesick et~al.(2023)Janesick, Shelansky, Gottscho, Wagner, Williams, Rouault, Beliakoff, Morrison, Oliveira, Sicherman, et~al.]{xenium}
Amanda Janesick, Robert Shelansky, Andrew~D Gottscho, Florian Wagner, Stephen~R Williams, Morgane Rouault, Ghezal Beliakoff, Carolyn~A Morrison, Michelli~F Oliveira, Jordan~T Sicherman, et~al.
\newblock High resolution mapping of the tumor microenvironment using integrated single-cell, spatial and in situ analysis.
\newblock \emph{Nature Communications}, 14\penalty0 (1):\penalty0 8353, 2023.

\bibitem[Jaume et~al.(2024{\natexlab{a}})Jaume, Doucet, Song, Lu, Almagro-Perez, Wagner, Vaidya, Chen, Williamson, Kim, and Mahmood]{hest}
Guillaume Jaume, Paul Doucet, Andrew~H. Song, Ming~Y. Lu, Cristina Almagro-Perez, Sophia~J. Wagner, Anurag~J. Vaidya, Richard~J. Chen, Drew F.~K. Williamson, Ahrong Kim, and Faisal Mahmood.
\newblock {HEST-1k: A Dataset for Spatial Transcriptomics and Histology Image Analysis}.
\newblock \emph{arXiv}, 2024{\natexlab{a}}.

\bibitem[Jaume et~al.(2024{\natexlab{b}})Jaume, Oldenburg, Vaidya, Chen, Williamson, Peeters, Song, and Mahmood]{tangle}
Guillaume Jaume, Lukas Oldenburg, Anurag~Jayant Vaidya, Richard~J. Chen, Drew~FK Williamson, Thomas Peeters, Andrew~H. Song, and Faisal Mahmood.
\newblock Transcriptomics-guided slide representation learning in computational pathology.
\newblock In \emph{CVPR}, 2024{\natexlab{b}}.

\bibitem[Jaume et~al.(2024{\natexlab{c}})Jaume, Vaidya, Chen, Williamson, Liang, and Mahmood]{jaume2023modeling}
Guillaume Jaume, Anurag Vaidya, Richard Chen, Drew Williamson, Paul Liang, and Faisal Mahmood.
\newblock Modeling dense multimodal interactions between biological pathways and histology for survival prediction.
\newblock \emph{CVPR}, 2024{\natexlab{c}}.

\bibitem[Jia et~al.(2021)Jia, Yang, Xia, Chen, Parekh, Pham, Le, Sung, Li, and Duerig]{align}
Chao Jia, Yinfei Yang, Ye Xia, Yi-Ting Chen, Zarana Parekh, Hieu Pham, Quoc Le, Yun-Hsuan Sung, Zhen Li, and Tom Duerig.
\newblock Scaling up visual and vision-language representation learning with noisy text supervision.
\newblock In \emph{ICML}, pages 4904--4916. PMLR, 2021.

\bibitem[Jia et~al.(2024)Jia, Liu, Chen, Zhao, and Wang]{thitogene}
Yuran Jia, Junliang Liu, Li Chen, Tianyi Zhao, and Yadong Wang.
\newblock Thitogene: a deep learning method for predicting spatial transcriptomics from histological images.
\newblock \emph{Briefings in Bioinformatics}, 25\penalty0 (1):\penalty0 bbad464, 2024.

\bibitem[Kueckelhaus et~al.(2024)Kueckelhaus, Frerich, Kada-Benotmane, Koupourtidou, Ninkovic, Dichgans, Beck, Schnell, and Heiland]{kueckelhaus2024inferring}
Jan Kueckelhaus, Simon Frerich, Jasim Kada-Benotmane, Christina Koupourtidou, Jovica Ninkovic, Martin Dichgans, Juergen Beck, Oliver Schnell, and Dieter~Henrik Heiland.
\newblock Inferring histology-associated gene expression gradients in spatial transcriptomic studies.
\newblock \emph{Nature Communications}, 15\penalty0 (1):\penalty0 7280, 2024.

\bibitem[Li et~al.(2021)Li, Li, and Eliceiri]{dsmil}
Bin Li, Yin Li, and Kevin~W Eliceiri.
\newblock Dual-stream multiple instance learning network for whole slide image classification with self-supervised contrastive learning.
\newblock In \emph{CVPR}, pages 14318--14328, 2021.

\bibitem[Li et~al.(2022)Li, Li, Xiong, and Hoi]{li2022blip}
Junnan Li, Dongxu Li, Caiming Xiong, and Steven Hoi.
\newblock Blip: Bootstrapping language-image pre-training for unified vision-language understanding and generation.
\newblock In \emph{ICML}, 2022.

\bibitem[Li and Wang(2021)]{li2021bulk}
Xinmin Li and Cun-Yu Wang.
\newblock From bulk, single-cell to spatial rna sequencing.
\newblock \emph{International journal of oral science}, 13\penalty0 (1):\penalty0 36, 2021.

\bibitem[Li et~al.(2024)Li, Ren, Huang, Zhao, Wang, Shen, Gao, Chen, Zhu, Xiong, et~al.]{li2024pibf1}
Xiaomin Li, Ci Ren, Anni Huang, Yue Zhao, Liming Wang, Hui Shen, Chun Gao, Bingxin Chen, Tong Zhu, Jinfeng Xiong, et~al.
\newblock Pibf1 regulates multiple gene expression via impeding long-range chromatin interaction to drive the malignant transformation of hpv16 integration epithelial cells.
\newblock \emph{Journal of Advanced Research}, 57:\penalty0 163--180, 2024.

\bibitem[Liu et~al.(2024)Liu, Liu, Xu, Cui, Ke, and Ma]{liu2024exploiting}
Mingxin Liu, Yunzan Liu, Pengbo Xu, Hui Cui, Jing Ke, and Jiquan Ma.
\newblock Exploiting geometric features via hierarchical graph pyramid transformer for cancer diagnosis using histopathological images.
\newblock \emph{IEEE Transactions on Medical Imaging}, 2024.

\bibitem[Lu et~al.(2021)Lu, Williamson, Chen, Chen, Barbieri, and Mahmood]{lu2021data}
Ming~Y Lu, Drew~FK Williamson, Tiffany~Y Chen, Richard~J Chen, Matteo Barbieri, and Faisal Mahmood.
\newblock Data-efficient and weakly supervised computational pathology on whole-slide images.
\newblock \emph{Nature biomedical engineering}, 5\penalty0 (6):\penalty0 555--570, 2021.

\bibitem[Lu et~al.(2024)Lu, Chen, Williamson, Chen, Liang, Ding, Jaume, Odintsov, Le, Gerber, et~al.]{conch}
Ming~Y Lu, Bowen Chen, Drew~FK Williamson, Richard~J Chen, Ivy Liang, Tong Ding, Guillaume Jaume, Igor Odintsov, Long~Phi Le, Georg Gerber, et~al.
\newblock A visual-language foundation model for computational pathology.
\newblock \emph{Nature Medicine}, 30:\penalty0 863–874, 2024.

\bibitem[Ludwig and Weinstein(2005)]{ludwig2005biomarkers}
Joseph~A Ludwig and John~N Weinstein.
\newblock Biomarkers in cancer staging, prognosis and treatment selection.
\newblock \emph{Nature Reviews Cancer}, 5\penalty0 (11):\penalty0 845--856, 2005.

\bibitem[Maynard et~al.(2021)Maynard, Collado-Torres, Weber, Uytingco, Barry, Williams, Catallini, Tran, Besich, Tippani, et~al.]{datasetdlpfc}
Kristen~R Maynard, Leonardo Collado-Torres, Lukas~M Weber, Cedric Uytingco, Brianna~K Barry, Stephen~R Williams, Joseph~L Catallini, Matthew~N Tran, Zachary Besich, Madhavi Tippani, et~al.
\newblock Transcriptome-scale spatial gene expression in the human dorsolateral prefrontal cortex.
\newblock \emph{Nature neuroscience}, 24\penalty0 (3):\penalty0 425--436, 2021.

\bibitem[Min et~al.(2024)Min, Shi, Zhang, Wan, and Wang]{mclstExp}
Wenwen Min, Zhiceng Shi, Jun Zhang, Jun Wan, and Changmiao Wang.
\newblock Multimodal contrastive learning for spatial gene expression prediction using histology images.
\newblock \emph{arXiv preprint arXiv:2407.08216}, 2024.

\bibitem[Oksza-Orzechowski et~al.(2024)Oksza-Orzechowski, Quinten, Shafighi, Kie{\l}basa, van Kessel, de~Groen, Vermaat, Sepl{\'u}veda-Y{\'a}{\~n}ez, Navarrete, Veelken, et~al.]{oksza2024caclust}
Kazimierz Oksza-Orzechowski, Edwin Quinten, Shadi~Darvish Shafighi, Szymon~M Kie{\l}basa, Hugo van Kessel, Ruben~AL de Groen, Joost~SP Vermaat, Julieta~H Sepl{\'u}veda-Y{\'a}{\~n}ez, Marcelo~A Navarrete, Hendrik Veelken, et~al.
\newblock Caclust: linking genotype to transcriptional heterogeneity of follicular lymphoma using bcr and exomic variants.
\newblock \emph{bioRxiv}, pages 2024--04, 2024.

\bibitem[Oord et~al.(2018)Oord, Li, and Vinyals]{oord2018representation}
Aaron van~den Oord, Yazhe Li, and Oriol Vinyals.
\newblock Representation learning with contrastive predictive coding.
\newblock \emph{arXiv preprint arXiv:1807.03748}, 2018.

\bibitem[Oquab et~al.(2023)Oquab, Darcet, Moutakanni, Vo, Szafraniec, Khalidov, Fernandez, Haziza, Massa, El-Nouby, Howes, Huang, Xu, Sharma, Li, Galuba, Rabbat, Assran, Ballas, Synnaeve, Misra, Jegou, Mairal, Labatut, Joulin, and Bojanowski]{oquab2023dinov2}
Maxime Oquab, Timothée Darcet, Theo Moutakanni, Huy~V. Vo, Marc Szafraniec, Vasil Khalidov, Pierre Fernandez, Daniel Haziza, Francisco Massa, Alaaeldin El-Nouby, Russell Howes, Po-Yao Huang, Hu Xu, Vasu Sharma, Shang-Wen Li, Wojciech Galuba, Mike Rabbat, Mido Assran, Nicolas Ballas, Gabriel Synnaeve, Ishan Misra, Herve Jegou, Julien Mairal, Patrick Labatut, Armand Joulin, and Piotr Bojanowski.
\newblock Dinov2: Learning robust visual features without supervision, 2023.

\bibitem[Pang et~al.(2021)Pang, Su, and Li]{histogene}
Minxing Pang, Kenong Su, and Mingyao Li.
\newblock Leveraging information in spatial transcriptomics to predict super-resolution gene expression from histology images in tumors.
\newblock \emph{BioRxiv}, pages 2021--11, 2021.

\bibitem[Qu et~al.(2024)Qu, Ma, Luo, Guo, Wang, and Song]{qu2024rethinking}
Linhao Qu, Yingfan Ma, Xiaoyuan Luo, Qinhao Guo, Manning Wang, and Zhijian Song.
\newblock Rethinking multiple instance learning for whole slide image classification: A good instance classifier is all you need.
\newblock \emph{IEEE Transactions on Circuits and Systems for Video Technology}, 2024.

\bibitem[Radford et~al.(2021)Radford, Kim, Hallacy, Ramesh, Goh, Agarwal, Sastry, Askell, Mishkin, Clark, et~al.]{clip}
Alec Radford, Jong~Wook Kim, Chris Hallacy, Aditya Ramesh, Gabriel Goh, Sandhini Agarwal, Girish Sastry, Amanda Askell, Pamela Mishkin, Jack Clark, et~al.
\newblock Learning transferable visual models from natural language supervision.
\newblock In \emph{ICML}, pages 8748--8763. PMLR, 2021.

\bibitem[Rousseeuw(1987)]{rousseeuw1987silhouettes}
Peter~J Rousseeuw.
\newblock Silhouettes: a graphical aid to the interpretation and validation of cluster analysis.
\newblock \emph{Journal of computational and applied mathematics}, 20:\penalty0 53--65, 1987.

\bibitem[Saillard et~al.(2024)Saillard, Jenatton, Llinares-López, Mariet, Cahané, Durand, and Vert]{hoptimus}
Charlie Saillard, Rodolphe Jenatton, Felipe Llinares-López, Zelda Mariet, David Cahané, Eric Durand, and Jean-Philippe Vert.
\newblock H-optimus-0, 2024.

\bibitem[Schaar et~al.(2024)Schaar, Tejada-Lapuerta, Palla, Gutgesell, Halle, Minaeva, Vornholz, Dony, Drummer, Bahrami, et~al.]{schaar2024nicheformer}
Anna~Christina Schaar, Alejandro Tejada-Lapuerta, Giovanni Palla, Robert Gutgesell, Lennard Halle, Mariia Minaeva, Larsen Vornholz, Leander Dony, Francesca Drummer, Mojtaba Bahrami, et~al.
\newblock Nicheformer: a foundation model for single-cell and spatial omics.
\newblock \emph{bioRxiv}, pages 2024--04, 2024.

\bibitem[Shao et~al.(2021)Shao, Bian, Chen, Wang, Zhang, Ji, et~al.]{transmil}
Zhuchen Shao, Hao Bian, Yang Chen, Yifeng Wang, Jian Zhang, Xiangyang Ji, et~al.
\newblock Transmil: Transformer based correlated multiple instance learning for whole slide image classification.
\newblock \emph{NIPS}, 34:\penalty0 2136--2147, 2021.

\bibitem[Shi et~al.(2023)Shi, Tang, Li, Zhang, Gao, Zheng, Wang, Gong, and Li]{shi2023structure}
Jiangbo Shi, Lufei Tang, Yang Li, Xianli Zhang, Zeyu Gao, Yefeng Zheng, Chunbao Wang, Tieliang Gong, and Chen Li.
\newblock A structure-aware hierarchical graph-based multiple instance learning framework for pt staging in histopathological image.
\newblock \emph{IEEE Transactions on Medical Imaging}, 42\penalty0 (10):\penalty0 3000--3011, 2023.

\bibitem[Shi et~al.(2024)Shi, Li, Gong, Zheng, and Fu]{shi2024vila}
Jiangbo Shi, Chen Li, Tieliang Gong, Yefeng Zheng, and Huazhu Fu.
\newblock Vila-mil: Dual-scale vision-language multiple instance learning for whole slide image classification.
\newblock In \emph{Proceedings of the IEEE/CVF Conference on Computer Vision and Pattern Recognition}, pages 11248--11258, 2024.

\bibitem[Song et~al.(2024)Song, Chen, Jaume, Vaidya, Baras, and Mahmood]{song2024multimodal}
Andrew~H Song, Richard~J Chen, Guillaume Jaume, Anurag~Jayant Vaidya, Alexander Baras, and Faisal Mahmood.
\newblock Multimodal prototyping for cancer survival prediction.
\newblock In \emph{ICML}, 2024.

\bibitem[Song et~al.(2020)Song, Chan, and Wei]{song2020flexible}
Fangda Song, Ga~Ming~Angus Chan, and Yingying Wei.
\newblock Flexible experimental designs for valid single-cell rna-sequencing experiments allowing batch effects correction.
\newblock \emph{Nature communications}, 11\penalty0 (1):\penalty0 3274, 2020.

\bibitem[St{\aa}hl et~al.(2016{\natexlab{a}})St{\aa}hl, Salm{\'e}n, Vickovic, Lundmark, Navarro, Magnusson, Giacomello, Asp, Westholm, Huss, et~al.]{staahl2016visualization}
Patrik~L St{\aa}hl, Fredrik Salm{\'e}n, Sanja Vickovic, Anna Lundmark, Jos{\'e}~Fern{\'a}ndez Navarro, Jens Magnusson, Stefania Giacomello, Michaela Asp, Jakub~O Westholm, Mikael Huss, et~al.
\newblock Visualization and analysis of gene expression in tissue sections by spatial transcriptomics.
\newblock \emph{Science}, 353\penalty0 (6294):\penalty0 78--82, 2016{\natexlab{a}}.

\bibitem[St{\aa}hl et~al.(2016{\natexlab{b}})St{\aa}hl, Salm{\'e}n, Vickovic, Lundmark, Navarro, Magnusson, Giacomello, Asp, Westholm, Huss, et~al.]{visium}
Patrik~L St{\aa}hl, Fredrik Salm{\'e}n, Sanja Vickovic, Anna Lundmark, Jos{\'e}~Fern{\'a}ndez Navarro, Jens Magnusson, Stefania Giacomello, Michaela Asp, Jakub~O Westholm, Mikael Huss, et~al.
\newblock Visualization and analysis of gene expression in tissue sections by spatial transcriptomics.
\newblock \emph{Science}, 353\penalty0 (6294):\penalty0 78--82, 2016{\natexlab{b}}.

\bibitem[Theodoris et~al.(2023)Theodoris, Xiao, Chopra, Chaffin, Al~Sayed, Hill, Mantineo, Brydon, Zeng, Liu, et~al.]{geneformer}
Christina~V Theodoris, Ling Xiao, Anant Chopra, Mark~D Chaffin, Zeina~R Al~Sayed, Matthew~C Hill, Helene Mantineo, Elizabeth~M Brydon, Zexian Zeng, X~Shirley Liu, et~al.
\newblock Transfer learning enables predictions in network biology.
\newblock \emph{Nature}, 618\penalty0 (7965):\penalty0 616--624, 2023.

\bibitem[Tian et~al.(2023)Tian, Chen, and Macosko]{tian2023expanding}
Luyi Tian, Fei Chen, and Evan~Z Macosko.
\newblock The expanding vistas of spatial transcriptomics.
\newblock \emph{Nature Biotechnology}, 41\penalty0 (6):\penalty0 773--782, 2023.

\bibitem[Van~der Maaten and Hinton(2008)]{van2008visualizingtsne}
Laurens Van~der Maaten and Geoffrey Hinton.
\newblock Visualizing data using t-sne.
\newblock \emph{Journal of machine learning research}, 9\penalty0 (11), 2008.

\bibitem[Vorontsov et~al.(2023)Vorontsov, Bozkurt, Casson, Shaikovski, Zelechowski, Liu, Severson, Zimmermann, Hall, Tenenholtz, et~al.]{vorontsov2023virchow}
Eugene Vorontsov, Alican Bozkurt, Adam Casson, George Shaikovski, Michal Zelechowski, Siqi Liu, Kristen Severson, Eric Zimmermann, James Hall, Neil Tenenholtz, et~al.
\newblock Virchow: a million-slide digital pathology foundation model.
\newblock \emph{arXiv preprint arXiv:2309.07778}, 2023.

\bibitem[Wang et~al.(2024{\natexlab{a}})Wang, Zhang, Zhu, Jiang, Qin, and Yuan]{wang2024mgiml}
Pengyu Wang, Huaqi Zhang, Meilu Zhu, Xi Jiang, Jing Qin, and Yixuan Yuan.
\newblock Mgiml: Cancer grading with incomplete radiology-pathology data via memory learning and gradient homogenization.
\newblock \emph{IEEE Transactions on Medical Imaging}, 2024{\natexlab{a}}.

\bibitem[Wang et~al.(2022)Wang, Yang, Zhang, Wang, Zhang, Yang, Huang, and Han]{wang2022transformer}
Xiyue Wang, Sen Yang, Jun Zhang, Minghui Wang, Jing Zhang, Wei Yang, Junzhou Huang, and Xiao Han.
\newblock Transformer-based unsupervised contrastive learning for histopathological image classification.
\newblock \emph{Medical image analysis}, 81:\penalty0 102559, 2022.

\bibitem[Wang et~al.(2024{\natexlab{b}})Wang, Zhao, Marostica, Yuan, Jin, Zhang, Li, Tang, Wang, Li, et~al.]{wang2024pathology}
Xiyue Wang, Junhan Zhao, Eliana Marostica, Wei Yuan, Jietian Jin, Jiayu Zhang, Ruijiang Li, Hongping Tang, Kanran Wang, Yu Li, et~al.
\newblock A pathology foundation model for cancer diagnosis and prognosis prediction.
\newblock \emph{Nature}, pages 1--9, 2024{\natexlab{b}}.

\bibitem[Wolf et~al.(2018)Wolf, Angerer, and Theis]{wolf2018scanpy}
F~Alexander Wolf, Philipp Angerer, and Fabian~J Theis.
\newblock Scanpy: large-scale single-cell gene expression data analysis.
\newblock \emph{Genome biology}, 19:\penalty0 1--5, 2018.

\bibitem[Xiao et~al.(2021)Xiao, Cong, Li, He, Wu, Tian, Wang, Yang, Liang, Liang, et~al.]{xiao2021cathepsin}
Yansen Xiao, Min Cong, Jiatao Li, Dasa He, Qiuyao Wu, Pu Tian, Yuan Wang, Shuaixi Yang, Chenxi Liang, Yajun Liang, et~al.
\newblock Cathepsin c promotes breast cancer lung metastasis by modulating neutrophil infiltration and neutrophil extracellular trap formation.
\newblock \emph{Cancer cell}, 39\penalty0 (3):\penalty0 423--437, 2021.

\bibitem[Xie et~al.(2024)Xie, Pang, Chung, Perciani, MacParland, Wang, and Bader]{bleep}
Ronald Xie, Kuan Pang, Sai Chung, Catia Perciani, Sonya MacParland, Bo Wang, and Gary Bader.
\newblock Spatially resolved gene expression prediction from histology images via bi-modal contrastive learning.
\newblock \emph{NIPS}, 36, 2024.

\bibitem[Xu et~al.(2024{\natexlab{a}})Xu, Fu, Long, Ang, Sethi, Chong, Li, Uddamvathanak, Lee, Ling, et~al.]{datasetbreast}
Hang Xu, Huazhu Fu, Yahui Long, Kok~Siong Ang, Raman Sethi, Kelvin Chong, Mengwei Li, Rom Uddamvathanak, Hong~Kai Lee, Jingjing Ling, et~al.
\newblock Unsupervised spatially embedded deep representation of spatial transcriptomics.
\newblock \emph{Genome Medicine}, 16\penalty0 (1):\penalty0 12, 2024{\natexlab{a}}.

\bibitem[Xu et~al.(2024{\natexlab{b}})Xu, Usuyama, Bagga, Zhang, Rao, Naumann, Wong, Gero, Gonz{\'a}lez, Gu, et~al.]{xu2024whole}
Hanwen Xu, Naoto Usuyama, Jaspreet Bagga, Sheng Zhang, Rajesh Rao, Tristan Naumann, Cliff Wong, Zelalem Gero, Javier Gonz{\'a}lez, Yu Gu, et~al.
\newblock A whole-slide foundation model for digital pathology from real-world data.
\newblock \emph{Nature}, pages 1--8, 2024{\natexlab{b}}.

\bibitem[Xu et~al.(2024{\natexlab{c}})Xu, Wang, Zhou, Ma, Yang, Lin, Wang, Wang, Liang, Han, et~al.]{xu2024multimodal}
Yingxue Xu, Yihui Wang, Fengtao Zhou, Jiabo Ma, Shu Yang, Huangjing Lin, Xin Wang, Jiguang Wang, Li Liang, Anjia Han, et~al.
\newblock A multimodal knowledge-enhanced whole-slide pathology foundation model.
\newblock \emph{arXiv preprint arXiv:2407.15362}, 2024{\natexlab{c}}.

\bibitem[Xu et~al.(2024{\natexlab{d}})Xu, Wang, Yang, Li, Ma, Chen, Wang, Huang, Gould, Lu, et~al.]{xu2024stomicsdb}
Zhicheng Xu, Weiwen Wang, Tao Yang, Ling Li, Xizheng Ma, Jing Chen, Jieyu Wang, Yan Huang, Joshua Gould, Huifang Lu, et~al.
\newblock Stomicsdb: a comprehensive database for spatial transcriptomics data sharing, analysis and visualization.
\newblock \emph{Nucleic acids research}, 52\penalty0 (D1):\penalty0 D1053--D1061, 2024{\natexlab{d}}.

\bibitem[Yu et~al.(2022)Yu, Wang, Vasudevan, Yeung, Seyedhosseini, and Wu]{yu2022coca}
Jiahui Yu, Zirui Wang, Vijay Vasudevan, Legg Yeung, Mojtaba Seyedhosseini, and Yonghui Wu.
\newblock Coca: Contrastive captioners are image-text foundation models.
\newblock \emph{arXiv preprint arXiv:2205.01917}, 2022.

\bibitem[Yu et~al.(2023)Yu, Xu, Zhang, and Li]{yu2023batch}
Xiaokang Yu, Xinyi Xu, Jingxiao Zhang, and Xiangjie Li.
\newblock Batch alignment of single-cell transcriptomics data using deep metric learning.
\newblock \emph{Nature communications}, 14\penalty0 (1):\penalty0 960, 2023.

\bibitem[Yuan et~al.(2024)Yuan, Ma, Gao, Cui, Wang, Fa, Ma, Wei, Ma, and Yu]{yuan2024heartsvg}
Xin Yuan, Yanran Ma, Ruitian Gao, Shuya Cui, Yifan Wang, Botao Fa, Shiyang Ma, Ting Wei, Shuangge Ma, and Zhangsheng Yu.
\newblock Heartsvg: a fast and accurate method for identifying spatially variable genes in large-scale spatial transcriptomics.
\newblock \emph{Nature Communications}, 15\penalty0 (1):\penalty0 5700, 2024.

\bibitem[Yuan et~al.(2023)Yuan, Pan, Zhao, Zhao, Xu, Li, Zhao, Zhang, and Yao]{yuan2023sodb}
Zhiyuan Yuan, Wentao Pan, Xuan Zhao, Fangyuan Zhao, Zhimeng Xu, Xiu Li, Yi Zhao, Michael~Q Zhang, and Jianhua Yao.
\newblock Sodb facilitates comprehensive exploration of spatial omics data.
\newblock \emph{Nature Methods}, 20\penalty0 (3):\penalty0 387--399, 2023.

\bibitem[Zeng et~al.(2022)Zeng, Wei, Yu, Yin, Yuan, Li, Tang, Lu, and Yang]{his2st}
Yuansong Zeng, Zhuoyi Wei, Weijiang Yu, Rui Yin, Yuchen Yuan, Bingling Li, Zhonghui Tang, Yutong Lu, and Yuedong Yang.
\newblock Spatial transcriptomics prediction from histology jointly through transformer and graph neural networks.
\newblock \emph{Briefings in Bioinformatics}, 23\penalty0 (5):\penalty0 bbac297, 2022.

\bibitem[Zhu and Thompson(2019)]{zhu2019metabolic}
Jiajun Zhu and Craig~B Thompson.
\newblock Metabolic regulation of cell growth and proliferation.
\newblock \emph{Nature reviews Molecular cell biology}, 20\penalty0 (7):\penalty0 436--450, 2019.

\end{thebibliography}
}

\clearpage \appendix \appendix
\renewcommand{\thefigure}{S\arabic{figure}}  
\renewcommand{\thetable}{S\arabic{table}}    
\setcounter{figure}{0}  
\setcounter{table}{0}   

\label{sec:appendix_section}
\section{More Experimental Results}
\subsection{Impact of Unimodal Pre-training}
\begin{figure}[tbp]
    \centering
    \includegraphics[width=0.5\textwidth]{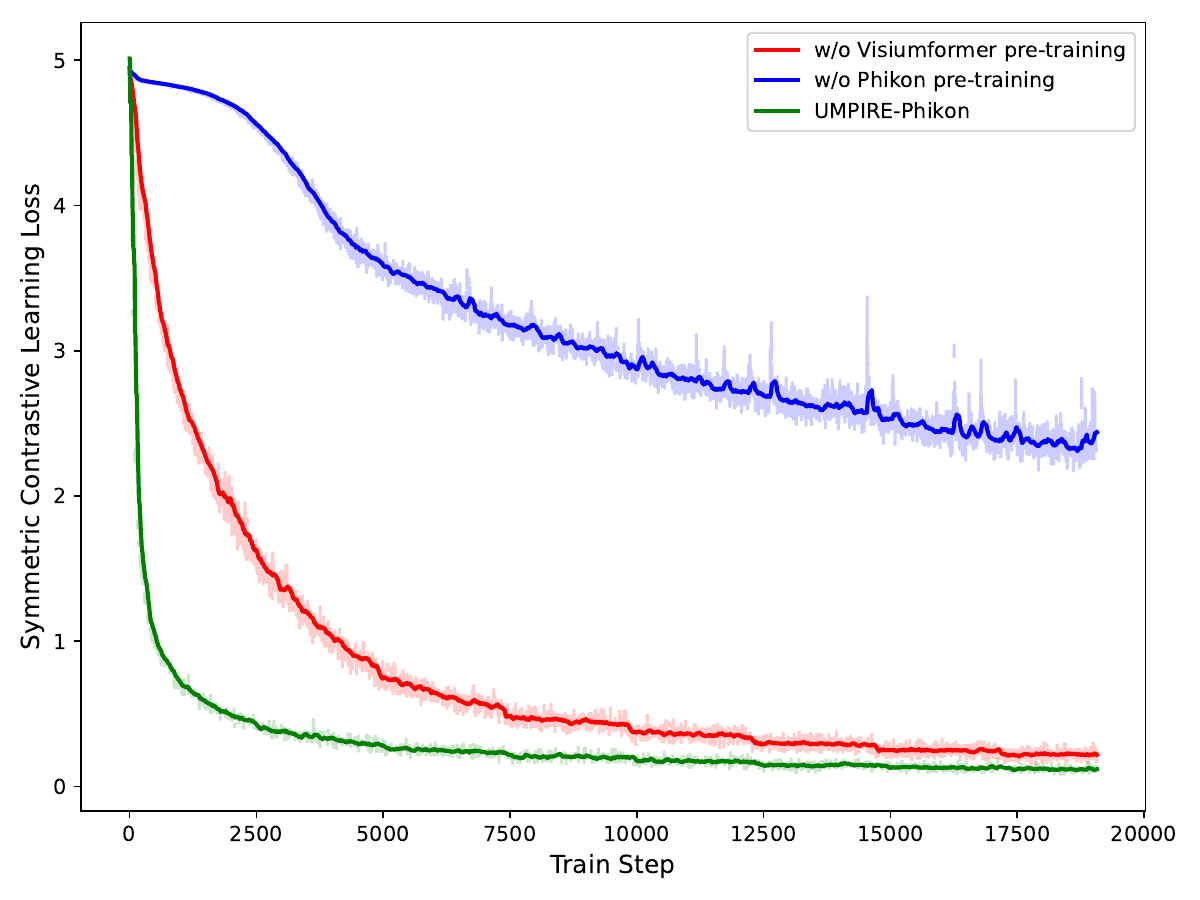}
    \caption{The impact of whether to conduct the first stage on the convergence speed of multimodal alignment pre-training.}
    \label{supp_fig:fig1}
\end{figure}

The pre-training process of $\ours$ is divided into two stages—unimodal encoder pre-training and multimodal alignment pre-training—to mitigate reliance on the quantity of paired pathology image-spatial transcriptomic gene data. Moreover, our experiments reveal that the first stage significantly accelerates the convergence of the loss function during the second stage. We conducted these experiments using $\ours$, which integrates Visiumformer and Phikon~\cite{filiot2023phikon}. Figure \ref{supp_fig:fig1} illustrates that the convergence speed of the model's loss is significantly impacted when Visiumformer skips the first-stage pre-training or when the pret-rained weights from Phikon are excluded.

\subsection{More Results of Multimodal Representation Learning}
\label{A.2}
\vspace{0.5pt}\noindent\textbf{The Results of Top 100 HEG and HVG Genes: } 
In Section \ref{sec:4.3}, the performance of $\ours$ and other methods were evaluated on the HLT, HPC, and HER2+ datasets by reporting the average Pearson correlation coefficient (PCC) for the top 50 highly expressed genes (HEG) and highly variable genes (HVG). Table \ref{supp_tab:table1} presents the PCC for the top 100 HEGs and HVGs across the three datasets, highlighting the consistent advantage of $\ours$. Interestingly, a decline in predictive performance is observed when transitioning from the top 50 to the top 100 genes, suggesting that the model is particularly adept at identifying patterns among genes with the highest expression levels or the greatest variability. This finding underscores the model's capacity to focus on genes that are more biologically significant and potentially more relevant in understanding complex biological processes. Furthermore, these genes are often the most informative markers of pathological alterations in tissues or tumors, emphasizing the practical utility of the approach for detecting critical molecular changes associated with disease states~\cite{yuan2024heartsvg,gu2023comprehensive,grun2020revealing}.

\begin{table*}
\centering
\resizebox{\textwidth}{!}{
\begin{tabular}{c|l|cccccc|c} 
\toprule
\multirow{2}{*}{Top 100}                                                            &\multicolumn{1}{c}{\multirow{2}{*}{Method}}  & \multicolumn{2}{c}{\textbf{HLT}}          & \multicolumn{2}{c}{\textbf{HPC}}          & \multicolumn{2}{c|}{\textbf{HER2+}} & \multirow{2}{*}{\textbf{Average}}  \\
                                                                                  &                         & HVG   & \multicolumn{1}{c|}{HEG} & HVG   & \multicolumn{1}{c|}{HEG} & HVG   & HEG                &                           \\ 
\midrule
\multirow{4}{*}{\begin{tabular}[c]{@{}c@{}}Regression\\based\end{tabular}}        & ST-Net~\cite{stnet}       & 0.0265$_{\pm0.0112}$  & 0.0301$_{\pm0.0076}$  & 0.1890$_{\pm0.1568}$    & 0.0631$_{\pm0.0480}$  & 0.1062$_{\pm0.0570}$ & 0.0940$_{\pm0.0413}$  & 0.0848                      \\
                                                                                  & HisToGene~\cite{histogene}& 0.0344$_{\pm0.0213}$  & 0.0387$_{\pm0.0284}$  & 0.1172$_{\pm0.0876}$    & 0.0888$_{\pm0.0387}$  & 0.0301$_{\pm0.0363}$  & 0.0228$_{\pm0.0299}$  & 0.0553                      \\
                                                                                  & His2ST~\cite{his2st}     &0.0051$_{\pm0.0125}$    & 0.0028$_{\pm0.0157}$  & 0.0224$_{\pm2.09}$      & 0.0138$_{\pm0.0129}$  & 0.0411$_{\pm0.0185}$  & 0.0298$_{\pm0.0177}$  & 0.0192                       \\
                                                                                  & THItoGene~\cite{thitogene}& 0.0055$_{\pm0.0124}$  & 0.0023$_{\pm0.0126}$  & 0.0311$_{\pm2.84}$      & 0.0193$_{\pm0.0246}$  & 0.0319$_{\pm0.0135}$  & 0.0207$_{\pm0.0098}$  & 0.0185                      \\ 
\midrule
\multirow{2}{*}{\begin{tabular}[c]{@{}c@{}}Contrastive learning\\based\end{tabular}} & mclSTExp~\cite{mclstExp}& 0.1530$_{\pm0.0313}$   & 0.2561$_{\pm0.0164}$ & 0.2738$_{\pm0.1272}$    & 0.0967$_{\pm0.0105}$  & 0.1324$_{\pm0.0713}$ & 0.0929$_{\pm0.0486}$ & 0.1675                    \\
                                                                                  & BLEEP~\cite{bleep}      & 0.1579$_{\pm0.0354}$   & 0.2530$_{\pm0.0195}$ & 0.2885$_{\pm0.1300}$    & 0.0999$_{\pm0.0432}$  & 0.1443$_{\pm0.0637}$ & 0.1283$_{\pm0.0562}$ & 0.1787                     \\ 
\midrule
\multirow{4}{*}{\begin{tabular}[c]{@{}c@{}}$\ours$-\textsc{Adapter}\\(Ours)\end{tabular}}  & \textit{Niche.} + Phikon&0.1478$_{\pm0.0383}$&0.2532$_{\pm0.0246}$&0.3630$_{\pm0.1604}$&0.1906$_{\pm0.0407}$    & 0.2329$_{\pm0.0881}$ & 0.2136$_{\pm0.0661}$ & 0.2335                         \\
                                                                                  & \textit{Niche.} + UNI   & 0.1559$_{\pm0.0365}$   & 0.2657$_{\pm0.0184}$ & 0.3896$_{\pm0.1481}$    & 0.1918$_{\pm0.0166}$ & 0.2409$_{\pm0.0872}$ & 0.2028$_{\pm0.0626}$ & 0.2412                          \\
                                                                                  & \textit{Visium.} + Phikon &0.1849$_{\pm0.0370}$  & \underline{0.2909}$_{\pm0.0193}$ & 0.3818$_{\pm0.1611}$    & 0.2114$_{\pm0.0376}$ & \textbf{0.2482}$_{\pm0.0846}$ & \textbf{0.2185}$_{\pm0.0635}$ & 0.2560                          \\
                                                                                  & \textit{Visium.} + UNI    &0.1854$_{\pm0.0371}$  & 0.2874$_{\pm0.0212}$ & 0.3781$_{\pm0.1580}$    & 0.1645$_{\pm0.0324}$ & \underline{0.2478}$_{\pm0.0898}$ & 0.2153$_{\pm0.0637}$ & 0.2464                          \\ 
\midrule
\multirow{6}{*}{\begin{tabular}[c]{@{}c@{}}$\ours$-\textsc{Finetune}\\(Ours)\end{tabular}} & \textit{Trans.} + Phikon  &0.1841$_{\pm0.0407}$ &0.2832$_{\pm0.0314}$ &0.3854$_{\pm0.1567}$ &0.2191$_{\pm0.0352}$&0.2048$_{\pm0.0858}$ &0.1674$_{\pm0.0621}$  &0.2407                           \\
                                                                                  & \textit{Trans.} + UNI     &0.1378$_{\pm0.0353}$    & 0.2252$_{\pm0.0194}$ & 0.3834$_{\pm0.1541}$    & 0.1941$_{\pm0.077}$  & 0.2069$_{\pm0.0802}$ & 0.1683$_{\pm0.0627}$ & 0.2193                          \\
                                                                                  & \textit{Niche.} + Phikon  & 0.1740$_{\pm0.0365}$   & 0.2680$_{\pm0.0212}$ & 0.3796$_{\pm0.1453}$    & 0.2069$_{\pm0.0227}$  & 0.2289$_{\pm0.0880}$ & 0.2023$_{\pm0.0620}$ & 0.2433                     \\
                                                                                  & \textit{Niche.} + UNI     & 0.1563$_{\pm0.0377}$   & 0.2588$_{\pm0.0236}$ & 0.3881$_{\pm0.1390}$    & 0.2146$_{\pm0.0317}$  & 0.2340$_{\pm0.0879}$ & 0.1983$_{\pm0.0651}$ & 0.2417                     \\
                                                                                  & \textit{Visium.} + Phikon & \underline{0.1855}$_{\pm0.0412}$   & 0.2838$_{\pm0.0249}$ & \textbf{0.3949}$_{\pm0.1483}$    & \textbf{0.2271}$_{\pm0.0281}$  & 0.2438$_{\pm0.0904}$ & 0.2175$_{\pm0.0681}$ & \underline{0.2588}                      \\
                                                                                  & \textit{Visium.} + UNI    & \textbf{0.1919}$_{\pm0.0368}$   & \textbf{0.2913}$_{\pm0.0246}$ & \underline{0.3898}$_{\pm0.1550}$    & \underline{0.2207}$_{\pm0.0296}$  & 0.2467$_{\pm0.0907}$& \underline{0.2177}$_{\pm0.0682}$ & \textbf{0.2597}                      \\
\bottomrule
\end{tabular}
}
\caption{
\textbf{Results of Gene Expression Prediction. }
The mean and standard deviation of the Pearson correlation coefficient (PCC) for the top 100 highly variable genes (HVG) and highly expressed genes (HEG). Where \textit{Visium.} refers to Visiumformer, \textit{Niche.} refers to Nicheformer, and \textit{Trans.} indicates a 12-layer Transformer without any pre-train. $\ours$-\textsc{Finetune} and $\ours$-\textsc{Adapter} represent full parameter fine-tuning and the use of adapter, respectively.
}
\label{supp_tab:table1}
\end{table*}

\begin{figure*}[htbp]
    \centering
    \includegraphics[width=\linewidth]{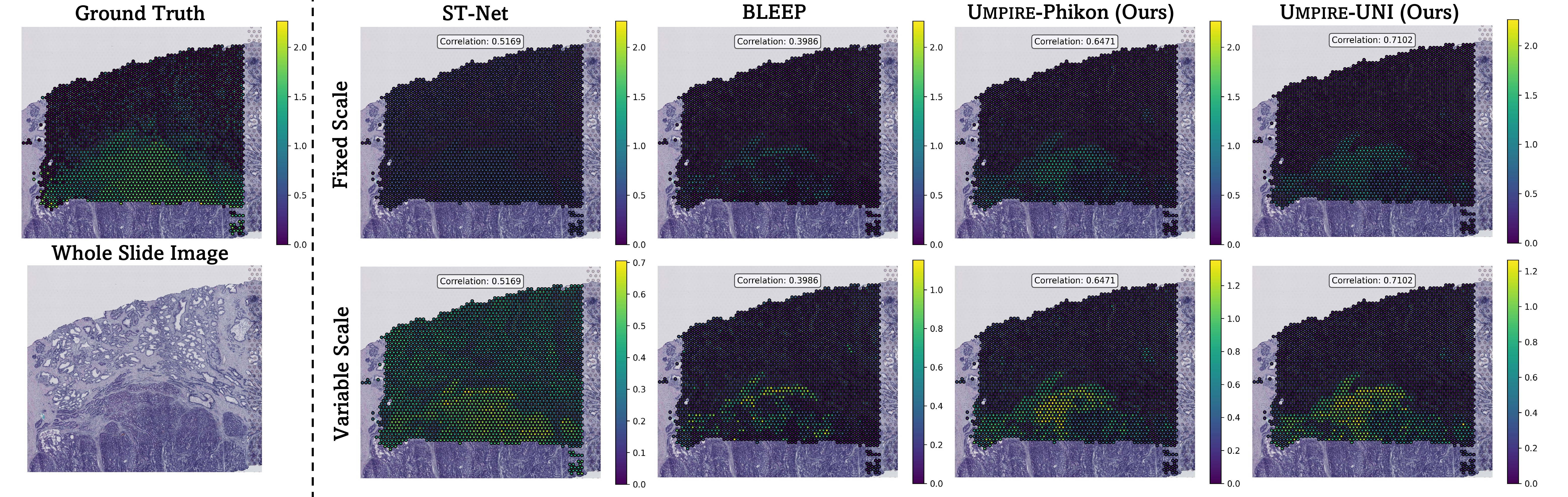}
    \vspace{-1em}
    \caption{
    \textbf{Visualization of bimodal-based gene expression prediction. } 
    Ground truth and predicted spatially resolved expression levels for CTSC overlaying the whole slide image of sample patient-1-H2-5, visualized with a fixed (top) and a variable (bottom) color scale.}
    \vspace{-1em}
    \label{supp_fig:fig2}
\end{figure*}

\begin{figure*}[htbp]
    \centering
    \includegraphics[width=\linewidth]{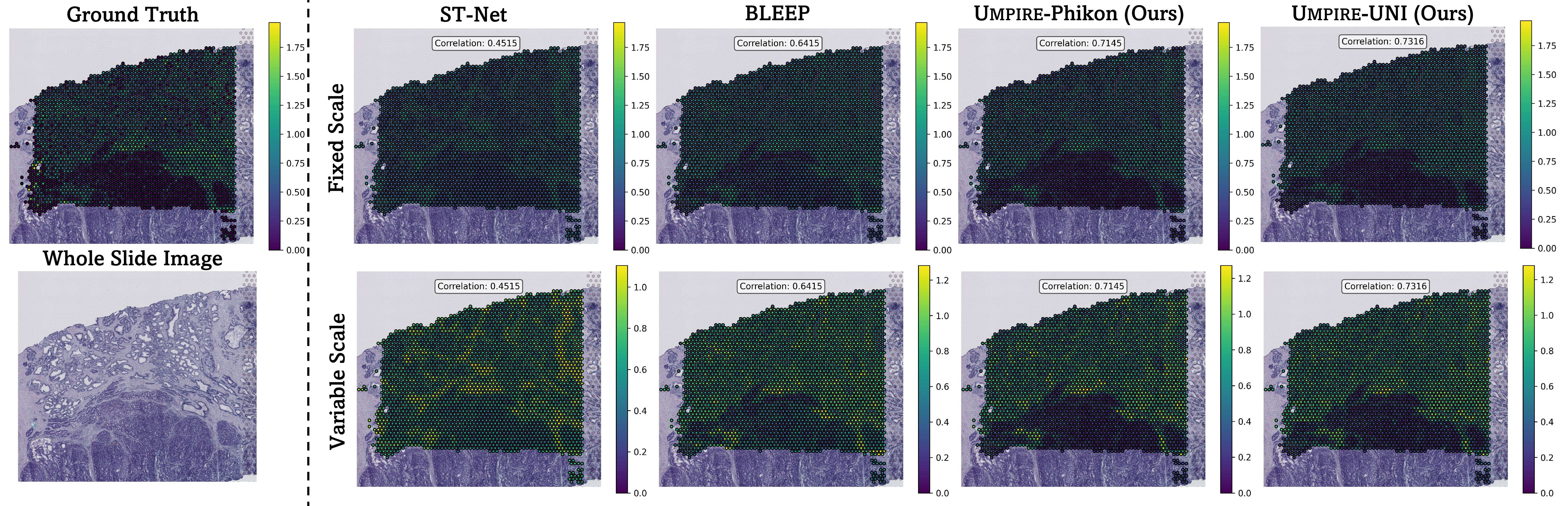}
    \vspace{-1em}
    \caption{
    \textbf{Visualization of bimodal-based gene expression prediction. } 
    Ground truth and predicted spatially resolved expression levels for H2AZ1 overlaying the whole slide image of sample patient-1-H2-5, visualized with a fixed (top) and a variable (bottom) color scale.}
    \label{supp_fig:fig3}
    \vspace{-1em}
\end{figure*}

\vspace{0.5pt}\noindent\textbf{Additional Case Study: }
In Section \ref{sec:4.3}, the predicted expression of the PIBF1 gene for sample patient-1-H2-5 was visualized using $\ours$ and other methods. Furthermore, the predicted expression levels were visualized alongside the ground truth for the CTSC (Figure \ref{supp_fig:fig2}) and H2AZ1 (Figure \ref{supp_fig:fig3}) genes for the same sample. Compared to other methods, $\ours$ demonstrates a superior ability to comprehensively preserve the heterogeneity of gene expression within tissue slices, particularly in distinguishing between tumor and normal regions. This enhanced capability allows clinicians and researchers to focus on areas of the tissue slices that provide greater informational value, thereby facilitating more targeted and insightful analyses.

\subsection{Zero-shot Embeddings Visualization}
\label{A.3}
\begin{figure*}[htbp]
    \centering
    \includegraphics[width=\linewidth]{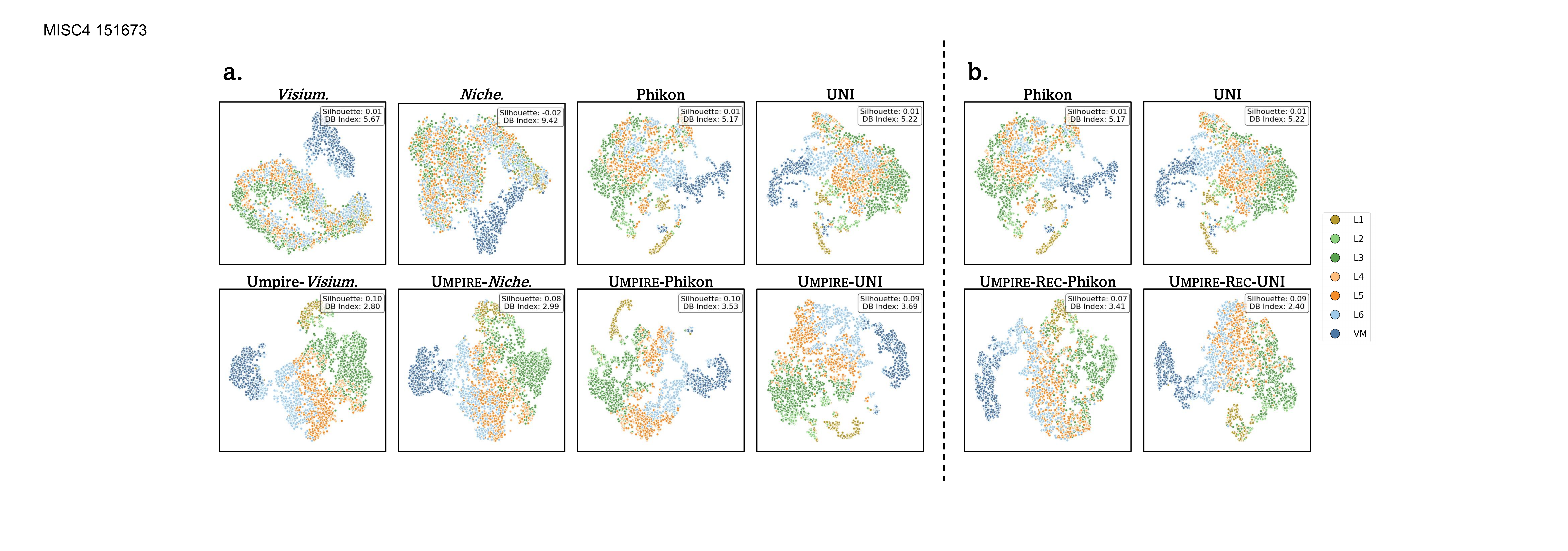}
    \vspace{-2em}
    \caption{
    \textbf{Classification visualization using t-SNE. } 
    \textbf{a.} t-SNE visualization of sample 151673 in the DLPFC dataset, visualized before (top) and after (bottom) multimodal pre-training with symmetric contrastive learning loss; 
    \textbf{b.} with reconstruction loss. We also report the Silhouette score (Silhouette, $\uparrow$)~\cite{rousseeuw1987silhouettes} and the Davies-Bouldin index (DB Index, $\downarrow$)~\cite{davies1979dbindex}.
    }
    \label{supp_fig:fig4}
    \vspace{-1.5em}
\end{figure*}

Following pre-training, we conducted a zero-shot t-Distributed Stochastic Neighbor Embedding (t-SNE)~\cite{van2008visualizingtsne} visualization on the DLPFC dataset, focusing on sample 151673, as shown in Figure \ref{supp_fig:fig4}. In addition, we evaluated the model's performance using the Silhouette score (Silhouette)~\cite{rousseeuw1987silhouettes} and the Davies-Bouldin Index (DB Index)~\cite{davies1979dbindex}. Prior to $\ours$ pre-training, the model could distinguish only the white matter (WM), with the remaining cortical layers (L1-L6) largely indistinguishable. Post-pre-training, however, the model exhibited a markedly improved capacity to differentiate among the cortical layers, accompanied by substantial improvements in both the Silhouette and DB Index, reflecting enhanced spatial and cluster separation.

\subsection{Loss Function Ablation Study}
\label{A.4}
\begin{figure}[tbp]
    \centering
    \includegraphics[width=\linewidth]{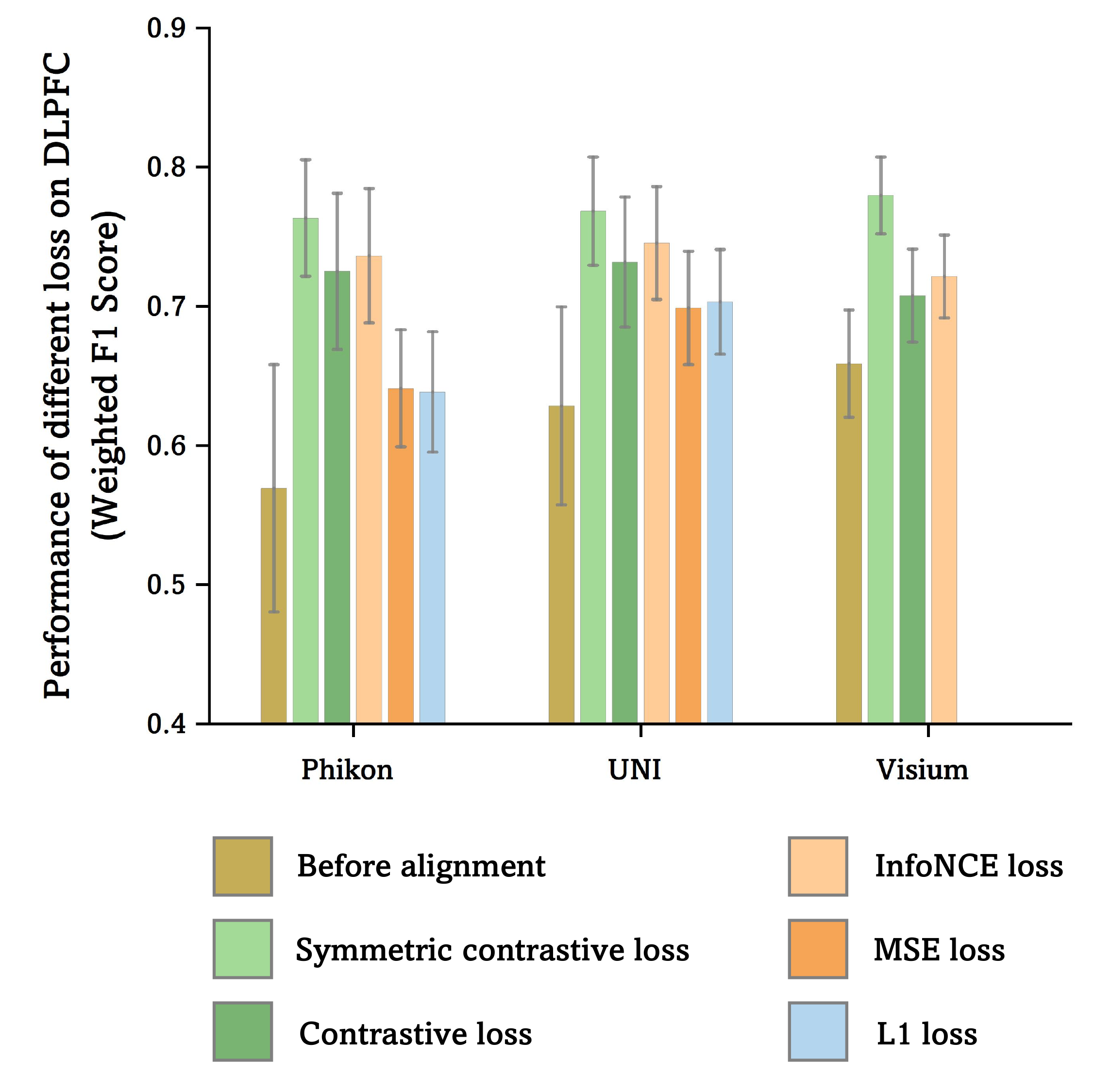}
    \caption{Loss function ablation study on DLPFC dataset.}
    \label{supp_fig:fig5}
\end{figure}

As detailed in Section \ref{sec:4.3}, we conducted an ablation study on the loss functions using the DLPFC dataset for the linear probing classification task. Replacing the symmetric contrastive loss (SCL) with reconstruction loss functions (mean squared error loss and L1 loss) resulted in a weighted F1 score reduction of $-16.0\%$ and $-16.4\%$, respectively, for Phikon. Similarly, substituting SCL with Contrastive loss~\cite{hadsell2006dimensionality} and InfoNCE loss~\cite{oord2018representation} led to weighted F1 score decreases of $-5.0\%$ and $-3.5\%$, respectively. The effects of different loss functions on $\ours$-UNI and $\ours$-\textit{Visium} were also analyzed, showing consistent performance degradation with loss function replacement. These findings are visualized in Figure \ref{supp_fig:fig5}. The superior performance of SCL can be attributed to the symmetry it introduces in contrastive learning, enabling more effective capture of bidirectional relationships within the data. This symmetry enhances the model's generalization across diverse tasks.

\subsection{Impact of Pathological Vision Encoder}
\label{A.5}
In the task of gene expression prediction, our model, $\ours$, achieved improvements in the PCC of $+215.8\%$ and $+42.9\%$ compared to ST-Net~\cite{stnet} and BLEEP~\cite{bleep}, respectively. $\ours$ employs UNI~\cite{chen2024uni} and Phikon~\cite{filiot2023phikon} as vision encoders to encode pathology images, whereas ST-Net and BLEEP utilize DenseNet-121~\cite{DenseNet} and ResNet-50~\cite{resnet} as their respective vision encoders. To demonstrate that the significant performance improvement of $\ours$ is not solely attributable to using more powerful pathology-specific vision encoders, we replaced the vision encoders in ST-Net and BLEEP with Phikon and conducted the same experiments. Table \ref{supp_tab:table2} reports the performance changes observed when the vision encoder in ST-Net was replaced. This modification led to performance improvements in the HLT and HPC datasets. However, a decline in performance was noted on the HER2+ dataset, suggesting dataset-specific effects of the encoder replacement. Overall, the performance of the modified ST-Net improved by $+32.1\%$ compared to the original ST-Net, yet it still significantly lagged behind that of the original BLEEP.
Conversely, the situation was entirely different for BLEEP; when we replaced the vision encoder in BLEEP with Phikon, the performance across all three datasets decreased, with an average decline of $-58.2\%$. This phenomenon has also been observed by Xie et al\@. \cite{bleep}, who attributed it to the use of large-parameter vision encoders on small-scale datasets. They argued that such an approach might lead the network to prioritize memorizing information within its weights rather than encoding it effectively in the projection space, ultimately compromising overall performance. In contrast, $\ours$ addresses this challenge by incorporating pre-training on extensive large-scale datasets, which enables the model to learn more robust and transferable representations. The improvements achieved by $\ours$ stem from the synergistic contributions of all modules and the strategic benefits of pre-training, rather than solely from replacing the vision encoder with a pathology-specific alternative.

\section{Model Architecture, Experiment Settings and Comparison Methods}
\label{B}
\subsection{Model Architecture}
\vspace{0.5pt}\noindent\textbf{Tokenization for Visiumformer: }
In biological experiments, systematic differences in measurement results, known as batch effects, can arise from variations in sample processing, experimental conditions, timing, operators, or other technical factors. These effects are particularly pronounced in high-throughput sequencing techniques, including RNA sequencing, single-cell sequencing, and spatial transcriptomics, and they can substantially influence data analysis and biological interpretation~\cite{song2020flexible,yu2023batch}. To mitigate batch effects, we standardized the count data across all spots, ensuring each spot contained 10,000 counts. Subsequently, we computed the average expression value for each gene across all data, considering only non-zero values in the calculation. The final normalized data were obtained by dividing the initial normalized values by the corresponding average expression values. The normalized results were then sorted in descending order, and the indices of the top $N$ genes were selected as the tokenized gene expression data. The complete normalization and tokenization procedure is detailed in Algorithm \ref{algorithm1}.

\vspace{0.5pt}\noindent\textbf{Model Architecture of Visiumformer: } 
Visiumformer is composed of 12 stacked Transformer blocks. As shown in Figure \ref{fig:fig1}, each Transformer block primarily consists of a multi-head attention mechanism and a feed-forward network (FFN). In this work, we use 16 attention heads, set the token dimension to $D=512$, and configure the hidden layer of the feed-forward network to 1024. For more details on the model architecture, please refer to Table \ref{supp_tab:table3}. 

\vspace{0.5pt}\noindent\textbf{Model Architecture of \textit{Trans.}: } 
To highlight the necessity of pre-training for Visiumformer, we designed a Transformer baseline model (\textit{Trans.}), described in Section \ref{sec:4.3}, where normalized gene expression values serve as input without any pre-training. The gene expression values for \textit{Trans.} were normalized using the same method as Visiumformer. Given the high dimensionality of gene expression data, the Scanpy library~\cite{wolf2018scanpy} was employed to select the top 1,500 highly variable genes across the training dataset. A $log1p$ transformation was then applied to prepare the input. This processed input was also used as the regression target for $\ours$-\textsc{Rec}. To ensure fairness in comparison, the Transformer blocks in \textit{Trans.} were kept identical to those in Visiumformer.

\subsection{Experiment Settings}
\label{B.2}
\begin{table*}[tp]
\centering
\resizebox{\textwidth}{!}{
\begin{tabular}{l|cccccc|c} 
\toprule
\multicolumn{1}{c|}{\multirow{2}{*}{\diagbox{Method}{PCC}}} & \multicolumn{2}{c}{\textbf{HLT}}              & \multicolumn{2}{c}{\textbf{HPC}}              & \multicolumn{2}{c|}{\textbf{HER2+}}     & \multirow{2}{*}{\textbf{Average}}  \\
\multicolumn{1}{c|}{}                                          & HVG                & \multicolumn{1}{c|}{HEG} & HVG                & \multicolumn{1}{c|}{HEG} & HVG                & HEG                &                                    \\ 
\midrule
ST-Net~\cite{stnet}                         & 0.0421$_{\pm0.0206}$  & 0.0406$_{\pm0.0140}$  & 0.2172$_{\pm0.1720}$    & 0.0445$_{\pm0.0386}$  & 0.1129$_{\pm0.0576}$  & 0.0940$_{\pm0.0421}$  & 0.0919                      \\
ST-Net-Phikon                               & 0.1090$_{\pm0.0294}$  & 0.1140$_{\pm0.0103}$  & 0.2326$_{\pm0.1557}$    & 0.1301$_{\pm0.0512}$  & 0.0842$_{\pm0.0597}$  & 0.0583$_{\pm0.0442}$  & 0.1214                      \\
\midrule
BLEEP~\cite{bleep}                          & 0.1995$_{\pm0.0435}$  & 0.2956$_{\pm0.0253}$  & 0.3221$_{\pm0.1417}$    & 0.0969$_{\pm0.0300}$  & 0.1692$_{\pm0.0729}$  & 0.1336$_{\pm0.0573}$  & 0.2028                     \\ 
BLEEP-Phikon                                & 0.0149$_{\pm0.0274}$  & 0.0240$_{\pm0.0334}$  & 0.2598$_{\pm0.1831}$    & 0.0786$_{\pm0.0481}$  & 0.0804$_{\pm0.0627}$  & 0.0513$_{\pm0.0493}$  & 0.0848                     \\ 
\bottomrule
\end{tabular}
}
\caption{
\textbf{Influence of Pathological Vision Encoder. }
The mean and standard deviation of the Pearson correlation coefficient (PCC) for the top 50 highly variable genes (HVG) and highly expressed genes (HEG). ST-Net-Phikon and BLEEP-Phikon denote the models in which Phikon has been substituted for the original vision encoders in the respective methods.}
\label{supp_tab:table2}
\end{table*}
\vspace{0.5pt}\noindent\textbf{Pre-training for Visiumformer: } 
Pre-training for Visiumformer was conducted using four NVIDIA A800 GPUs. The configurations for this per-training, including hyperparameters and setup details, are thoroughly outlined in Table \ref{supp_tab:table3}.

\begin{algorithm*}[tbp]
\SetKwInOut{Input}{Input}
\SetKwInOut{Output}{Output}
\Input{$ \mathbf{mean} \in \mathbb{R}^{20310}$: average expression across all data \\
       $ \mathbf{raw} \in \mathbb{R}^{B \times 20310}$: original gene expression \\
       $ N \in \mathbb{Z}$: number of contextual tokens.}
\Output{$\mathbf{T} \in \mathbb{R}^{B \times N}$: tokenized gene expression.}

$\mathbf{raw} \gets \text{ReplaceNaN}(\mathbf{raw}, 0)$ \tcp*[r]{Replace NaN values in $\mathbf{raw}$ with 0}

\For{$i \gets 0$ \KwTo $B - 1$}{
    $c_i \gets \sum_{j=1}^{20310} \mathbf{raw}[i, j]$ \tcp*[r]{Sum across rows}
    $c_i \gets c_i + (c_i == 0)$  \tcp*[r]{Avoid division by zero}
    $\mathbf{raw}[i] \gets \mathbf{raw}[i] \times \frac{10000}{c_i}$ \tcp*[r]{Normalize to 10000 counts}
    $\mathbf{raw}[i] \gets \mathbf{raw}[i] \oslash \mathbf{mean}$ \tcp*[r]{Mitigation batch effect}
    $\mathbf{T}[i] \gets \text{argsort}(\mathbf{raw}[i], \text{descending})[:N]$ \tcp*[r]{Select top $N$ tokens by descending order}
}
\caption{Tokenization of Raw Gene Expression}
\label{algorithm1}
\end{algorithm*}

\vspace{0.5pt}\noindent\textbf{Pre-training for Alignment: } 
All pre-training experiments for alignment were conducted using four NVIDIA A800 GPUs. Additional experimental configurations are provided in Table \ref{supp_tab:table4}. In addition, gene expression hidden states were extracted from the $12$-th Transformer block, and mean pooling was applied across the sequence length dimension to obtain the encoded gene expression embedding.

\vspace{0.5pt}\noindent\textbf{Experimental Platform for Downstream Tasks: } 
We evaluated our $\ours$ on multiple downstream tasks, all of which were performed on a single NVIDIA A800 GPU. 

\vspace{0.5pt}\noindent\textbf{Experiment Settings for Multimodal Representation Learning: } 
When fine-tuning on downstream datasets, leave-one-out cross-validation was employed, using one slice as the test set, while the remaining slices were used for training and validation. The model architecture is kept identical to that during pre-training. We set the learning rate to 1e-4, weight decay to 1e-3, and did not use warmup. AdamW was used as the optimizer. In addition, $80\%$ of the training data is used as the training set, and the remaining $20\%$ is used as the validation set. All models were trained for 50 epochs, with early stopping based on the validation loss and a patience of 5. When implementing $\ours$-\textsc{Adapter}, two linear layers with ReLU activation were incorporated following the Gene Encoder and the Vision Encoder, with a bottleneck layer dimension set to 128.

\begin{table}[tbp]
\centering
\resizebox{0.48\textwidth}{!}{
\begin{tabular}{cll}
\toprule
\multicolumn{1}{l}{} &
\textbf{Hyperparameter} & \textbf{Value} \\
\midrule
\multirow{9}{*}{\rotatebox[origin=c]{90}{Model Architecture}} & Vocab size & 20,310 \\
 & Token dimensionality & 512 \\
 & FFN dimensionality & 1024 \\
 & Number of Transformer layers & 12 \\
 & Max sequence length & 1,500 \\
 & Number of attention heads & 16 \\
 & Dropout & 0.0 \\
 & Hidden act & ReLU \\
 & LayerNorm eps & 1e-12 \\
\midrule
\multirow{9}{*}{\rotatebox[origin=c]{90}{Training Details}} & Optimizer & AdamW \\
 & Scheduler & CosineWarmupScheduler \\
 & Max learning rate & 1e-4 \\
 & Min learning rate & 1e-5 \\
 & Warm up steps & 20,000 \\
 & Total steps & 1,000,000 \\
 & Weight decay & 0.1 \\
 & Global batch size & 256 \\
 & Masking probability & 0.15 \\
\bottomrule
\end{tabular}}
\caption{Experiment Configurations for Visiumformer Pretrain.}
\label{supp_tab:table3}
\end{table}

\begin{table}[tbp]
\centering
\begin{tabular}{ll} 
\toprule
\textbf{Hyperparameter} & \textbf{Values}        \\
\midrule
Similarity function     & Cosine similarity      \\
Optimizer               & AdamW                  \\
Scheduler               & CosineWarmupScheduler  \\
Max learning rate       & 1e-4                   \\
Min learning rate       & 1e-5                   \\
Warm up steps           & 5,000                   \\
Total epochs            & 10                     \\
Weight decay            & 1e-3                   \\
Globa batch size        & 512                    \\
Extraction layer        & 12                     \\
Pooling method          & Mean                   \\
\bottomrule
\end{tabular}
\caption{Experiment Configurations for Alignment Pretrain.}
\label{supp_tab:table4}
\end{table}

\vspace{0.5pt}\noindent\textbf{Experiment Settings for Linear Probing: } 
Since the DLPFC dataset consists of 12 slices, we similarly employed leave-one-out cross-validation. In contrast, the 10X Breast dataset contains only a single slice, so five-fold cross-validation was used for this dataset. Adam was selected as the optimizer, with the learning rate set to 1e-4. The feature encoders were frozen, and only a trainable linear layer was added. All models were trained for 50 epochs, configuring early stopping with a patience 5. 

\vspace{0.5pt}\noindent\textbf{Experiment Settings for MIL-based WSI Classification: } 
We used CLAM~\cite{lu2021data} to divide all WSIs into non-overlapping patches of $256 \times 256$ pixels at 20$\times$ magnification. To meet the input requirements of the vision encoder, all patches were resized to $224 \times 224$ pixels. Since each patient may have multiple WSIs, five-fold cross-validation was performed at the patient level to prevent data leakage. When a patient had multiple WSIs, the patches obtained from all WSIs were stacked into a single bag. The simple yet effective ABMIL framework~\cite{ilse2018attention} was utilized as the feature aggregation module, while the cross-entropy loss was employed to guide the training process. All models were set with a learning rate of 5e-4, used Adam as the optimizer, and were trained for 50 epochs with early stopping and a patience of 5. 

\subsection{Downstream Comparison Methods}
To comprehensively evaluate the capabilities of $\ours$, in Section \ref{sec:4.3}, we compared several models, including regression-based models: ST-Net~\cite{stnet}, HisToGene~\cite{histogene}, His2ST~\cite{his2st}, and THItoGene~\cite{thitogene}, as well as contrastive learning-based models: BLEEP~\cite{bleep}, and mclSTExp~\cite{mclstExp}. 

\vspace{0.5pt}\noindent\textbf{ST-Net} is a deep learning model developed to integrate spatial transcriptomics data with pathology images for predicting gene expression in breast cancer. The model processes hematoxylin and eosin (H\&E)-stained tissue image patches of $224 \times 224$ pixels, corresponding to spots approximately 100 µm in diameter. It utilizes DenseNet-121~\cite{DenseNet} to extract image features, followed by a fully connected layer to predict the expression levels of 250 target genes. We only modified the fully connected layer to enable it to predict the full-dimensional gene expression.

\vspace{0.5pt}\noindent\textbf{HisToGene} utilizes a modified Vision Transformer architecture to account for the spatial dependencies between spatial transcriptomics spots. It first extracts image patches corresponding to the spatial coordinates of each spot in the spatial transcriptomics data. These patches are then processed through a learnable linear layer to generate patch embeddings and positional embeddings to capture spatial relationships. HisToGene employs multi-head attention layers to model these dependencies and predict gene expression.

\vspace{0.5pt}\noindent\textbf{His2ST} integrates Convolutional Neural Networks (CNNs) and Graph Convolutional Networks (GCNs) to predict spatial gene expression from histopathological images. CNNs are used to extract local features from the input images, capturing the tissue's morphological characteristics. GCNs then model the spatial relationships between neighbouring regions, enabling the model to effectively capture the spatial dependencies of gene expression within the tissue.

\vspace{0.5pt}\noindent\textbf{THItoGene} integrates dynamic convolutional networks, Efficient Capsule Networks, Vision Transformers, and Graph Attention Networks. By synthesizing these advanced components, THItoGene effectively captures local visual features, spatial dependencies, and inter-spot relationships. This powerful combination enables accurate high-resolution gene expression prediction from pathology images.

\vspace{0.5pt}\noindent\textbf{BLEEP} is a framework that utilizes contrastive learning to predict gene expression from pathology images. The model learns a joint low-dimensional embedding space from paired pathology images and gene expression profiles. Given a query image patch, BLEEP imputes gene expression by referencing the nearest neighbours in the learned embedding space from a reference dataset. This framework enables accurate and efficient prediction of spatially resolved gene expression profiles, outperforming existing methods in terms of prediction accuracy while preserving biological heterogeneity and robustness to experimental artifacts.

\vspace{0.5pt}\noindent\textbf{mclSTExp} employs a Transformer-based architecture to explicitly model spatial dependencies in spatial transcriptomics. It treats spatial transcriptomics spots as ``words” in a sequence, utilizing self-attention mechanisms to integrate positional and contextual information. By incorporating image features via contrastive learning, mclSTExp improves the accuracy of spatial gene expression predictions, especially in capturing complex tissue structures.

\section{Complexity Analysis}
\subsection{Complexity Analysis of Visiumformer}
Visiumformer is built on the Transformer and BERT architectures, which means that its time and space complexity bottleneck arises from the self-attention mechanism, characterized by a complexity of $O(N_g^2 \times L_g \times d_g)$, where $N_g$ represents the context length of the tokens input into Visiumformer, $L_g$ denotes the number of Transformer blocks, and $d_g$ denotes the embedding dimension.

\begin{table}[tbp]
\centering
\resizebox{0.48\textwidth}{!}{
\begin{tabular}{c|l|ccc} 
\toprule
Top 50                                                                            & \multicolumn{1}{c|}{\textbf{Method}} & \begin{tabular}[c]{@{}c@{}}\textbf{Trainable}\\\textbf{Param. ($\downarrow$)}\end{tabular} & \begin{tabular}[c]{@{}c@{}}\textbf{Training}\\\textbf{FLOPs ($\downarrow$)}\end{tabular} & \begin{tabular}[c]{@{}c@{}}\textbf{Average}\\\textbf{PCC ($\uparrow$)}\end{tabular}  \\ 
\midrule
\multirow{4}{*}{\begin{tabular}[c]{@{}c@{}}Regression\\based\end{tabular}}        & ST-Net~\cite{stnet}                              & 27.77M                                                                      & \underline{17.27G}                                                         & 0.0848                                                                  \\
                                                                                  & HisToGene~\cite{histogene}                           & 242.35M                                                                     & \textbf{1.45G}                                                         & 0.0553                                                                  \\
                                                                                  & His2ST~\cite{his2st}                              & 92.34M                                                                      & 108.3G                                                                 & 0.0192                                                                  \\
                                                                                  & THItoGene~\cite{thitogene}                           & 83.60M                                                                      & 82.11G                                                                 & 0.0185                                                                  \\ 
\midrule
\multirow{2}{*}{\begin{tabular}[c]{@{}c@{}}Contrastive learning\\based\end{tabular}} & mclSTExp~\cite{mclstExp}                            & 23.21M                                                                      & 17.24G                                                                 & 0.1675                                                                  \\
                                                                                  & BLEEP~\cite{bleep}                               & 24.55M                                                                      & 24.63G                                                                 & 0.1787                                                                  \\ 
\midrule
\multirow{2}{*}{\begin{tabular}[c]{@{}c@{}}$\ours$-Adapter\\(Ours)\end{tabular}}  & \textit{Visium.} + Phikon            & \textbf{0.92M}                                                              & 105.46G                                                                & 0.2560                                                                  \\
                                                                                  & \textit{Visium.} + UNI               & \underline{1.05M}                                                               & 358.09G                                                                & 0.2464                                                                  \\ 
\midrule
\multirow{2}{*}{\begin{tabular}[c]{@{}c@{}}$\ours$-Finetune\\(Ours)\end{tabular}} & \textit{Visium.} + Phikon            & 135.76M                                                                     & 332.23G                                                                & \underline{0.2588}                                                          \\
                                                                                  & \textit{Visium.} + UNI               & 353.44M                                                                     & 584.86G                                                                & \textbf{0.2597}                                                         \\
\bottomrule
\end{tabular}
}
\caption{\textbf{Complexity Analysis.}
Trainable Parameters, Training FLOPs, and Average PCC for $\ours$-\textsc{Finetune}, $\ours$-\textsc{Adapter}, and Comparative Methods.
}
\label{supp_tab:table5}
\end{table}

\subsection{Complexity Analysis of \textbf{$\ours$}}
$\ours$ primarily consists of two branches: the gene encoder, Visiumformer, and the vision encoder, ViT. Therefore, its time complexity is $O(N_g^2 \times L_g \times d_g + N_h^2 \times L_h \times d_h)$, where $N_h$ represents the context length of the vision encoder, $L_h$ denotes the number of Transformer blocks, and $d_h$ denotes the embedding dimension. For a batch of data, the time complexity of training $\ours$ can be expressed as: $O(B \times (N_g^2 \times L_g \times d_g + N_h^2 \times L_h \times d_h) + B^2)$, where $B$ represents the batch size and $ B^2 $ represents the complexity involved in computing the symmetric contrastive learning loss. In the task of gene expression prediction using multimodal representation learning, the time complexity of $\ours$ for inferring a single image is $O(N_g^2 \times L_g \times d_g + M \times d)$, where $M$ denotes the number of reference embeddings, and $d$ represents the dimensionality of the aligned embeddings.

\subsection{Comparison with Baseline Methods}
In Table \ref{supp_tab:table5}, the trainable parameters, training FLOPs, and average PCC for $\ours$-\textsc{Finetune}, $\ours$-\textsc{Adapter}, and other comparative methods are reported. Due to the limitations of previous methods, which were constrained to single, independent small datasets, there was a tendency to utilize simpler model architectures to mitigate overfitting. In contrast, our approach benefits from extensive pre-training on large-scale datasets followed by fine-tuning on downstream datasets. As a result, $\ours$ is capable of employing more complex and powerful vision and gene encoders without the risk of overfitting. Using more complex models has significantly increased the computational complexity of $\ours$. However, given the substantial performance improvement that accompanies this increase, we believe that this trade-off is acceptable.
Furthermore, we have developed a more efficient fine-tuning method called $\ours$-\textsc{Adapter}. This approach reduces the trainable parameters to just $0.7\%$ and $0.3\%$ of their original values. Similarly, the computational complexity is lowered to $31.7\%$ and $61.2\%$ of the initial levels, while the performance experiences only an average decrease of $-2.8\%$.

\begin{figure*}[tbp]
    \centering
    \includegraphics[width=\linewidth]{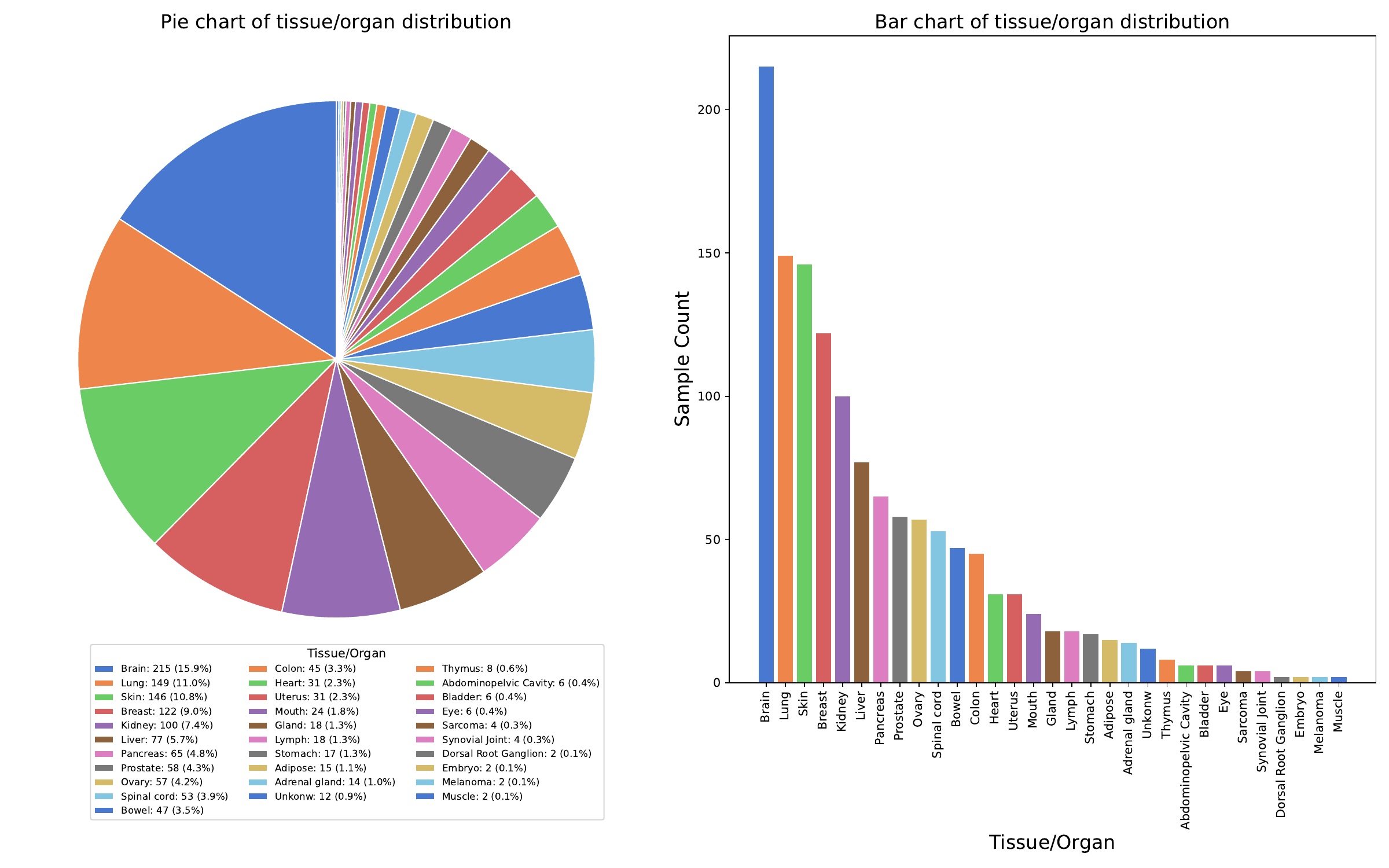}
    \vspace{-2em}
    \caption{Distribution of Organs or Tissues in the ViSTomics-4M Dataset.}
    \label{supp_fig:fig6}
\end{figure*}

\section{Datasets}
\subsection{ViSTomics-4M}
\label{D.1}
For the pre-training of Visiumformer, several of the largest existing datasets were combined, including SpatialOmics (55 slices)~\cite{yuan2023sodb}, STOmicsDB (302 slices)~\cite{xu2024stomicsdb}, HEST (308 slices)~\cite{hest}, and STimage-1K4M (309 slices)~\cite{chen2024stimage}. In addition, we downloaded human spatial transcriptomics data generated using Visium technology from the Gene Expression Omnibus  (389 slices), forming what is currently the largest \textbf{Vis}ium-based \textbf{S}patial \textbf{T}ranscript\textbf{omics} Dataset (ViSTomics-4M). ViSTomics-4M consists of 3.94 million spatial transcriptomics gene expression entries from 1,363 slices and 180 datasets or publications. As shown in Figure \ref{supp_fig:fig6}, ViSTomics-4M includes spatial transcriptomics data from 30 different tissues and organs, such as the brain, lungs, skin, and breast. This diversity ensures that Visiumformer can comprehensively learn the contextual information expressed in human spatial transcriptomic gene expression. 

\subsection{Data for Alignment}
\label{data alignment}
The HEST dataset~\cite{hest} was filtered to retain only the human data generated using the Visium platform. Additionally, only those spots located within tissues were kept. Spots with fewer than 100 detected gene expression were removed as well. Since certain data from HEST will be used in downstream tasks, these data are excluded during the pre-training phase to prevent any potential data leakage. For the pathology images, $224 \times 224$ pixels patches were extracted from the original WSIs based on the centre point coordinates. This resulted in most patches covering a distance of 50-100 micrometres ($\mu m$), sufficient to encompass the corresponding spot (diameter of 55 $\mu m$). After these processing steps, there are 696,636 pairs of pathology images and spatial transcriptomic gene expression available for alignment pre-training, sourced from 329 slices across 16 different tissues or organs. 

\subsection{More Information about Downstream Datasets}
\vspace{0.5pt}\noindent\textbf{HLT:} The Human Liver Tissue dataset~\cite{datasetpsc} (HLT) consists of four tissue sections from one healthy individual, resulting in a total of 9,254 paired histological images and gene expression. The HLT measurements were conducted using the Visium platform, and the four slices used for the experiments are named C73-A1-VISIUM, C73-B1-VISIUM, C73-C1-VISIUM, and C73-D1-VISIUM. After the quality control mentioned in Section \ref{data alignment}, these four slices retained 2,377, 2,342, 2,275, and 2,260 pairs of histological images and gene expression, respectively.

\vspace{0.5pt}\noindent\textbf{HPC:} The Human Prostate Cancer dataset~\cite{datasethpc} (HPC) consists of 37 sections from two prostate cancer patients. Five sections were selected from those patients, resulting in 14,783 paired samples. The HPC measurements were also conducted using the Visium platform, and the five slices used for the experiments are named patient-2-V2-2, patient-2-H2-2, patient-1-H2-5, patient-1-H2-2, and patient-1-H2-1. After quality control, these five slices retained 3,749, 3,047, 2,698, 2,781, and 2,500 pairs of histological images and gene expression, respectively. 

\vspace{0.5pt}\noindent\textbf{HER2+:} The HER2-positive breast tumor dataset~\cite{datasether2+} (HER2+) consists of 36 sections from eight patients. Following ST-Net~\cite{stnet}, we reserved 32 slides from seven patients, resulting in 11,509 data pairs. Unlike HLT and HPC, HER2+ was measured using the Spatial Transcriptomics platform~\cite{staahl2016visualization}. Notably, the model did not include any data based on Spatial Transcriptomics technology during the pre-training phase. The purpose of adding the HER2+ dataset is to assess the generalization capability of the $\ours$ across different sequencing technologies.

\vspace{0.5pt}\noindent\textbf{DLPFC:} The human dorsolateral prefrontal cortex dataset (DLPFC)~\cite{datasetdlpfc} comprises 12 sections from three healthy donors. Each spot was categorized into seven classes: white matter (WM) and layers L1–L6, resulting in 47,329 data pairs. DLPFC was measured using the Visium platform.

\vspace{0.5pt}\noindent\textbf{10X Breast:} The Human Breast Cancer dataset (10X Breast) comprises a single section from an invasive ductal carcinoma, with each spot classified into four categories: Surrounding Tumor, Invasive, Healthy, and Tumor~\cite{datasetbreast}. This results in a total of 3,789 paired data points. The dataset was generated using the Visium platform. 

\vspace{0.5pt}\noindent\textbf{LUAD-mutation:} The LUAD-mutation dataset consists of 692 Fresh Frozen WSIs from 437 patients in TCGA-LUAD. Following DeepPATH~\cite{coudray2018classification}, we aim to predict the WSI mutation state (positive/negative) in four specific genes: EGFR, KRAS, STK11, and TP53.

\section{Evaluation Metric}
\vspace{0.5pt}\noindent\textbf{PCC:} Pearson correlation coefficient (PCC) is a statistical measure that quantifies the strength and direction of a linear relationship between two quantitative variables. It is widely used in statistics to assess how closely two variables are related. The PCC ranges from $-1$ to $+1$; the larger the value, the more similar the two variables are. The formula for calculating $PCC_i$ for gene $i$ can be expressed as follows: 
\begin{align}
PCC_i=\frac{Cov(\mathbf{ep}_i,\hat{\mathbf{ep}}_i)}{Var(\mathbf{ep}_i)\times Var(\hat{\mathbf{ep}}_i)},
\end{align}
where $Cov(\cdot)$ represents the covariance, $Var(\cdot)$ denotes the variance, $\mathbf{ep}_i$ and $\hat{\mathbf{ep}}_i$ represent the ground truth and predicted values of gene $i$ across the entire slice, respectively. 

\vspace{0.5pt}\noindent\textbf{Silhouette Score:} The Silhouette Score~\cite{rousseeuw1987silhouettes} is a widely used metric for evaluating the quality of clusters produced by clustering algorithms. It provides a quantitative measure of how well-defined and distinct the clusters are, allowing researchers to assess the effectiveness of their clustering results. The Silhouette Score quantifies the cohesion and separation of data points within clusters. It ranges from $-1$ to $+1$. The larger the value, the better the clustering effect. The Silhouette Score for a single data point $ i $ is calculated using the following formula:
\begin{align}
Silhouette\ Score &= \frac{1}{N} \sum_{i=1}^{N} \frac{b_i - a_i}{\max(a_i, b_i)},
\end{align}
where $ a_i $ represents the average distance from a data point $ i $ to all other points within the same cluster, referred to as the intra-cluster distance. Conversely, $ b_i $ denotes the average distance from point $ i $ to all points in the nearest neighbouring cluster, known as the inter-cluster distance. And $N$ is the total number of data points.

\vspace{0.5pt}\noindent\textbf{Davies-Bouldin Index: }The Davies-Bouldin Index (DB Index) quantifies the average similarity between each cluster and its most similar counterpart.  Specifically, a DB Index close to zero suggests that clusters are well-separated and compact, while a higher DB Index indicates that the clusters are overlapping or poorly defined.

The DB Index is calculated by first determining the centroid of each cluster as the mean of its points. The intra-cluster distance, \( S_k \), is then computed as the average distance between the points and the centroid:

\begin{align}
S_k = \frac{1}{|C_k|} \sum_{i \in C_k} d(i, \mu_k),
\end{align}
where $ d(i, \mu_k) $ is the distance between point $ i $ and the centroid $ \mu_k $, and $ |C_k| $ is the number of points in cluster $ C_k $. Next, the inter-cluster distance for each pair of clusters is computed as \( M_{ij} = d(\mu_i, \mu_j) \), where \( \mu_i \) and \( \mu_j \) are the centroids. The DB Index is then derived by averaging the maximum similarity ratio for each cluster relative to all others:

\begin{align}
DB\ Index = \frac{1}{k} \sum_{i=1}^{k} \max_{j \neq i} \left( \frac{S_i + S_j}{M_{ij}} \right),
\end{align}
where $ S_i $ and $ S_j $ are the intra-cluster distances for clusters $ i $ and $ j $, and $ k $ represents the total number of clusters.

\section{Limitations and Widespread Social Impact}
\vspace{0.5pt}\noindent\textbf{Limitations: }Despite our efforts to collect as much data as possible for training $\ours$, the dataset remains relatively limited compared to those used in mainstream multimodal contrastive learning~\cite{clip,conch}. This limitation arises from the high costs of spatial transcriptomics, privacy concerns related to patient data, substantial heterogeneity among sequencing platforms, and the inherent interdisciplinary challenges. While visual encoders for pathology have been extensively studied, robust gene expression encoders tailored to spatial transcriptomics are still lacking. In this work, we trained the Visiumformer; however, we recognize that the gene encoder’s performance remains suboptimal, largely due to its simplistic framework and design. This limitation highlights significant opportunities for enhancing its capacity to effectively encode gene expression data. Additionally, although we validated the transferability between the Visium~\cite{visium} and Spatial Transcriptomics~\cite{staahl2016visualization} platforms, data from other spatial transcriptomics technologies were not included due to challenges in data acquisition. Future research should prioritize the following directions: 1) assembling larger and more diverse datasets; 2) training advanced gene expression encoders specific to spatial transcriptomics; and 3) developing multimodal models capable of robust generalization across different platforms and technologies.

\vspace{0.5pt}\noindent\textbf{Widespread Social Impact: }
Technological advancements should benefit a broader population. While molecular-level analyses of cancer significantly enhance diagnostic accuracy and subsequent precision treatments, the prohibitive costs of genomic sequencing and spatial transcriptomics currently limit these technologies to a select few. Our mission is to advance efficient and cost-effective pathological data analysis methods that incorporate molecular perspectives, thereby supporting cancer research and providing particular assistance to underserved regions. Meanwhile, computational pathology and spatial transcriptomics are evolving at an unprecedented pace. However, due to inherent challenges, the intersection and collaboration between these two fields remain in their infancy. Our efforts are dedicated to facilitating a more profound and comprehensive integration between these disciplines, thereby nurturing a synergistic environment that propels advancements at an accelerated pace.

\end{document}